\documentclass{article}

\PassOptionsToPackage{numbers, compress}{natbib}

\usepackage{iclr2025_conference,times}

\usepackage[utf8]{inputenc} 
\usepackage[T1]{fontenc}    
\usepackage{url}            
\usepackage{booktabs}       
\usepackage{amsfonts}       
\usepackage{nicefrac}       
\usepackage{microtype}      
\usepackage{xcolor}         
\usepackage[pagebackref=true,breaklinks=true,colorlinks=true,bookmarks=false,linkcolor={red!50!black},urlcolor={darkgray!80!black},citecolor={green!50!black}]{hyperref}

\usepackage{changes}

\newcommand{\figref}[1]{Fig.~\ref{#1}}
\newcommand{\tabref}[1]{Tab.~\ref{#1}}
\newcommand{\secref}[1]{Sec.~\ref{#1}}

\newcommand{\equref}[1]{Equ. (\ref{#1})}

\usepackage{amsmath}
\usepackage{overpic}
\usepackage{fontawesome}
\usepackage{booktabs}
\usepackage{amssymb}
\usepackage{bbding}
\usepackage{pifont}
\usepackage{wasysym}
\usepackage{utfsym}

\usepackage{arydshln}

\usepackage{xcolor}
\usepackage{enumitem}
\usepackage{xspace}
\usepackage{booktabs}
\usepackage{colortbl}
\usepackage{multirow}
\usepackage{makecell}
\usepackage{lipsum} 
\usepackage{gensymb} 
\usepackage{graphicx}
\usepackage{amssymb}
\usepackage{wrapfig}
\usepackage{bm}

\definecolor{MyDarkRed}{rgb}{0.66, 0.16, 0.16}
\definecolor{MyDarkBlue}{rgb}{0.16, 0.16, 0.66}

\usepackage{booktabs}   
\usepackage{subcaption} 

\usepackage{algorithm}
\usepackage{xcolor}
\usepackage[noend]{algpseudocode}
\usepackage{algorithmicx}
\usepackage{appendix}
\usepackage{multirow}
\usepackage{diagbox}

\newcommand{\eref}[1]{Eq.~\eqref{#1}}

\makeatletter
\DeclareRobustCommand\onedot{\futurelet\@let@token\@onedot}
\def\@onedot{\ifx\@let@token.\else.\null\fi\xspace}

\makeatother

\usepackage[labelsep=period]{caption}
\renewcommand{\paragraph}[1]{\vspace{0.05cm}\noindent \textbf{#1 \hspace{0.2em}}}

\iclrfinalcopy 

\definecolor{mygray}{gray}{.92}

\title{Flow Distillation Sampling: Regularizing 3D Gaussians with Pre-trained Matching Priors}

\author{%
  Lin-Zhuo Chen \thanks{Equal contribution.}  \\
  Nanjing University
  \And
   Kangjie Liu $^{*}$ \\
  Nanjing University \\
  \And
   Youtian Lin \\
  Nanjing University \\
  \And
  Siyu Zhu\\
  Fudan University \\
  \And
  Zhihao Li\\
  Huawei Noah’s Ark Lab \\
  \And
  Xun Cao\\
  Nanjing University \\
  \And
  Yao Yao \thanks{Corresponding author.} \\
  Nanjing University \\
}

\begin{document}
\maketitle
\begin{abstract}

3D Gaussian Splatting (3DGS) has achieved excellent rendering quality with fast training and rendering speed. However, its optimization process lacks explicit geometric constraints, leading to suboptimal geometric reconstruction in regions with sparse or no observational input views.
In this work, we try to mitigate the issue by incorporating a pre-trained matching prior to the 3DGS optimization process. We introduce Flow Distillation Sampling (FDS), a technique that leverages pre-trained geometric knowledge to bolster the accuracy of the Gaussian radiance field.
Our method employs a strategic sampling technique to target unobserved views adjacent to the input views, utilizing the optical flow calculated from the matching model (Prior Flow) to guide the flow analytically calculated from the 3DGS geometry (Radiance Flow).
Comprehensive experiments in depth rendering, mesh reconstruction, and novel view synthesis showcase the significant advantages of FDS over state-of-the-art methods. 
Additionally, our interpretive experiments and analysis aim to shed light on the effects of FDS on geometric accuracy and rendering quality, potentially providing readers with insights into its performance. Project page: \url{https://nju-3dv.github.io/projects/fds} .

\end{abstract}
\section{Introduction}
\label{sec:intro}

3D Gaussian Splatting (3DGS)~\citep{kerbl20233d} has been widely applied to the field of 3D reconstruction and rendering, including novel view synthesis of static scenes~\citep{kerbl20233d, yu2024mip}, mesh surface reconstruction~\citep{guedon2024sugar, yu2024gaussian}, inverse rendering~\citep{liang2024gs, gao2023relightable}, and dynamic 3D reconstruction~\citep{wu20244d, lin2024gaussian},
However, in scenarios with less-observed areas, such as indoor scenes and unbounded scenes, radiance field optimization often suffers from overfitting to these limited input views~\citep{li2024dngaussian}, 
resulting in unreliable and corrupted 
geometry reconstruction.

To mitigate the issue, recent research efforts
~\citep{li2024dngaussian, paliwal2024coherentgs, turkulainen2024dnsplatter} 
have focused on incorporating geometric priors 
from input views into the training process, thereby 
regulating the optimization of radiance field 
represented by 3D Gaussian points. 
For instance, DN-Splatter~\citep{turkulainen2024dnsplatter} 
integrates sensor depth and normal cues 
into the reconstruction process. 
However, sensor depth acquisition is costly,
and the depth prior information from pre-trained 
monocular deep models inevitably suffer from the scale ambiguity~\citep{liu2023robust}.
While the normal prior provides even better geometric details, the scale ambiguity still exists due to its monocular nature.



%
In contrast to monocular priors, pairwise matching priors can provide absolute scale information of the scene. 
In this paper, we introduce \textbf{F}low \textbf{D}istillation \textbf{S}ampling (FDS), an online method for distilling matching prior from a pre-trained optical flow model into the 3DGS training process. FDS aims to enhance the geometry quality of Gaussian radiance field by leveraging the matching prior into the unobserved novel view. 
Specifically, we observe that the flow between the input view and the unobserved view generated by the match prior model (i.e., \textbf{Prior Flow}), can guide and refine the flow analytically calculated from the 3DGS geometry (i.e., \textbf{Radiance Flow}), improving the 3DGS reconstruction quality. 
Moreover, better 3DGS scene will lead to more accurate Prior Flow, creating a mutually reinforcing effect between two computed flow maps. This remains effective even when the radiance field is poorly optimized and the image rendered from the unobserved viewpoint is blurry during training. 
In addition, a camera sampling scheme is proposed to adaptively control the overlap between input view and sampled view for better Prior Flow calculation, which allows to leverage prior geometric knowledge more profoundly and thereby better enhance the 3DGS reconstruction quality.


The proposed FDS has been extensively evaluated on MushRoom~\citep{ren2024mushroom}, ScanNet (V2)~\citep{dai2017scannet}, and Replica~\citep{replica19arxiv} datasets for the task of geometry reconstruction.
We apply FDS to two commonly used baseline approaches, namely 3DGS~\citep{kerbl20233d} and 2DGS ~\citep{Huang2DGS2024}.
The results demonstrate a significant improvement in geometry reconstruction accuracy.
Additionally, we provide interpretive experiments and comprehensive analysis on the effectiveness of FDS, shedding light on its influence on the quality of geometric reconstruction and novel-view rendering.
Our contributions are summarized as follows:
\begin{itemize}
    \item FDS leverages matching prior information to recover absolute scale, 
    significantly enhancing the geometric quality of the Gaussian radiance field.
    \item An adaptive camera sampling scheme is proposed to selectively choose unobserved views with controllable overlap with the input view, further improving the geometric quality of the Gaussian radiance field even in less-observed areas.
    
    \item By employing FDS and the adaptive camera sampling scheme, 
    we bring significant improvements to state-of-the-art 3D geometry reconstruction approaches.
\end{itemize}

\begin{figure} 
  \centering
  \begin{overpic}[width=1.\columnwidth,trim=40 140 20 90,clip]{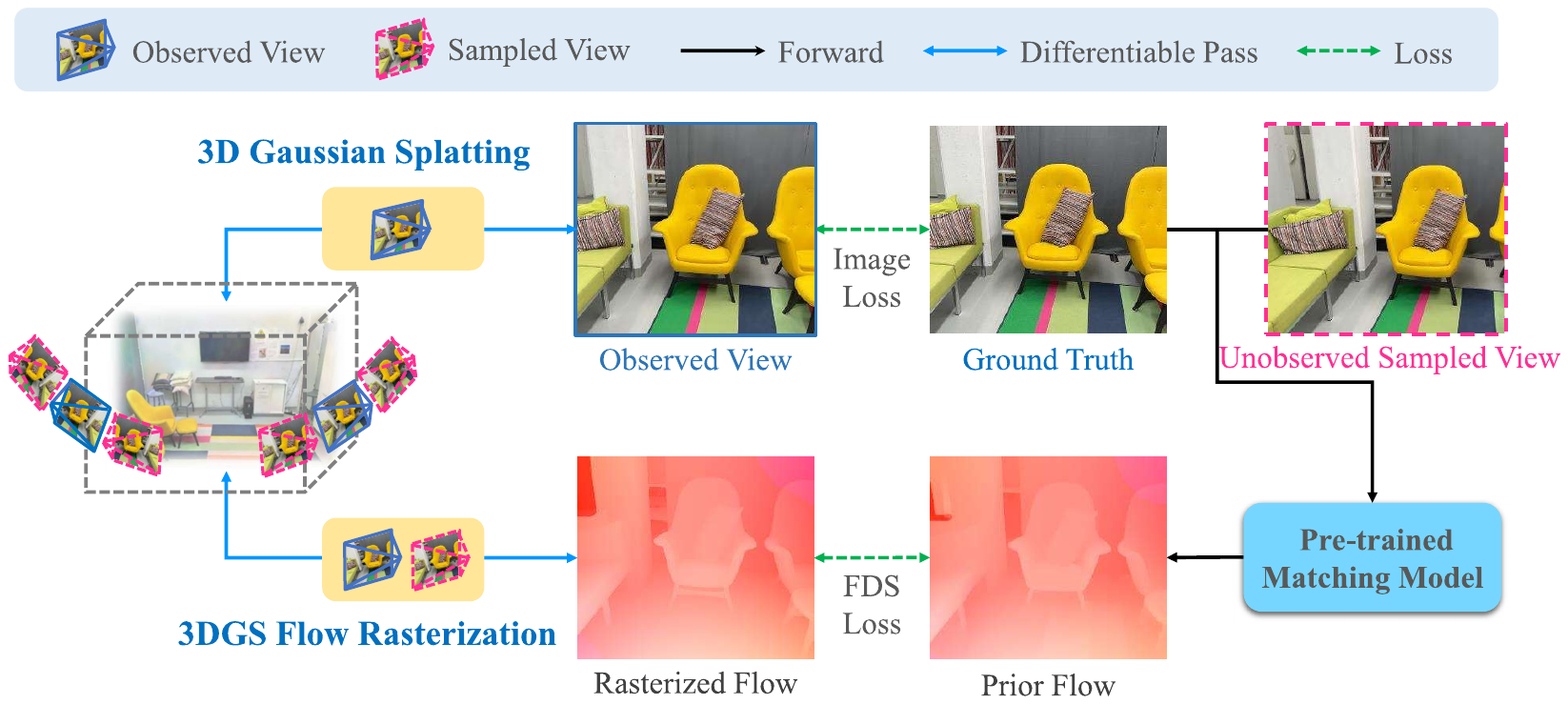}
	\end{overpic}
  \caption{\textbf{Pipeline of the proposed FDS.} For each
  input view, we apply the FDS camera sampling scheme to generate corresponding
  unobserved sampled view. We then compute Radiance flow base on rendered depth and
  the Prior flow from matching prior model. Finally the Prior Flow is used to supervise Radiance flow, which enhances the geometric quality of Gaussian Radiance Field.}
  \label{fig:fig1}
\end{figure}

\section{Related Work}
\label{sec:relatedwork}

\textbf{Geometry Reconstruction Based on 3DGS}: 
To enhance the geometry of 3DGS representations, 
some studies~\citep{gao2024mesh, waczynska2024games, lyu20243dgsr, chen2023neusg} 
focus on integrating mesh or SDF representations with 3DGS.
~\citep{gao2024mesh} improve the splitting of 3DGS under the guidance of
mesh representation and deformation gradients. 
A hybrid representation that integrates 3DGS with mesh, 
allowing 3DGS to be modified as a mesh, is introduced in 
~\citep{waczynska2024games}.
Mani-GS~\citep{gao2024mani} binds Gaussian and shape-aware triangular mesh to manipulate 3DGS directly.
3DGSR~\citep{lyu20243dgsr} combines a signed distance field (SDF) 
network with 3DGS to enhance geometry quality, 
aligning the SDF network’s geometry with that of 3DGS.
NeuSG~\citep{chen2023neusg} utilizes point clouds from 
3DGS to regulate NeuS, while its normals are also used 
to refine 3DGS. Additionally, NeuSG includes regularizers 
to ensure that 3DGS remains close to the surface.
Some work~\citep{guedon2023sugar, huang20242d, yu2024gaussian, chen2024pgsr} 
tends to extract geometry from 3DGS directly, benefiting from 
its fast training and rendering speed.
Sugar ~\citep{guedon2023sugar}
proposes an efficient mesh extraction method from 3DGS, aligned with
a regularization term during the training process and a refinement strategy.
2DGS~\citep{huang20242d} proposed a 2D surface modeling and two regularization 
losses which can preserve perspective-correct splatting and enchance geometry 
reconstruction.
A ray-tracing-based volume rendering is introduced in 
GOF~\citep{yu2024gaussian},
allowing the extraction of geometry from 3DGS directly.
PGSR~\citep{chen2024pgsr} proposes unbiased depth rendering
and single\& multi-view regularization loss to preserve 
geometric consistency.
However, in low-texture and less frequently observed areas, 
such as indoor scenes, 3DGS still tends to overfit to 
the limited input views, necessitating regulation 
through additional prior cues.

\textbf{Prior Regulation for Rendering}:
Prior information is usually incoporated in ill-posed problems,
including sparse view novel view synthesis, dynamic scene reconstruction,
and mesh reconstruction.
For sparse novel view synthesis tasks, most works utilize the prior depth information~\citep{deng2022depth, roessle2022dense, song2024darf, wang2023sparsenerf, zhu2023fsgs, xiong2023sparsegs, paliwal2024coherentgs}, semantic information~\citep{jain2021putting, wynn2023diffusionerf, xiong2023sparsegs}, and matching information~\citep{paliwal2024coherentgs, lao2024corresnerf} 
to constrain the optimization process of 3DGS.
Dynamic scene reconstruction is another challenging
task which requires reconstructing the scene geometry and object motion at the same
time. 
Therefore, optical flow priors are crucial as
they can help distinguish between camera motion 
and object motion~\citep{liu2023robust} 
while providing motion priors between frames~\citep{Liu_2023_CVPR, gao2021dynamic, li2023dynibar, 
wang2023flow, guo2023forward, tian2023mononerf}.
For the mesh reconstruction task,
geometry quality is also
enhanced by depth priors  
~\citep{wei2021nerfingmvs, yu2022monosdf, turkulainen2024dn} 
or normal priors ~\citep{yu2022monosdf, turkulainen2024dn}. 
However, using optical flow model priors to obtain metric depth priors 
to help geometric reconstruction, 
while leveraging unobserved regions to enhance the quality 
of limited view reconstruction, has not yet been explored.

\section{Method}
\label{sec:method}

Our FDS regulates the optimization 
of Gaussian radiance field by incorporating 
matching priors from the pretrained deep model.
The generation of Radiance Flow and
our proposed FDS loss, along with the equipped camera sampling scheme,  
are detailed in 
\secref{sec:method:subsec:gs} and \secref{sec:method:fds}, 
respectively.

\subsection{3D Gaussian Splatting and Radiance Flow}
\label{sec:method:subsec:gs}

We utilize 3DGS as an example to demonstrate
how Radiance Flow is generated.
3DGS~\citep{kerbl20233d} employs a 
collection of 3D Gaussians to represent 
the 3D scene. 
The expression of $i$-th 3D Gaussian distribution shows below:

\begin{equation}
    \mathcal{G}_i(\bm{X}) 
    = e^{-\frac{1}{2} (\bm{X} - \bm{\mu_i})^T \bm{\Sigma_i} (\bm{X} - \bm{\mu_i})},
\end{equation}

where $\bm{\Sigma_i} = \bm{R_iS_i}\bm{S_i}^T\bm{R_i}^T$, 
and $\bm{\mu_i}$ represents the position of the Gaussian, which is optimized during training.
Both $\bm{S_i}$ and $\bm{R_i}$ are represented by 
a 3D scaling vector $\bm{s_i}$ and a quaternion $\bm{q_i}$, respectively.
In addition to the above parameters, each Gaussian $\bm{P_i}$ has extra learnable attributes, 
including opacity $\bm{\alpha_i}$ and color
feature $\bm{f_i}$.
To summarize, $\bm{P_i} = \{\bm{\mu_i}. \bm{s_i}, \bm{q_i}, \bm{\alpha_i}, \bm{f_i}\}$.

To render a color for pixel $x$, 
a volume rendering based method similar to NeRF~\citep{mildenhall2021nerf}
is adapted:

\begin{equation}
    \bm{C}(\bm{x}) = \sum_{i \in N} \bm{c_i} \bm{\hat{\alpha_i}} \prod_{j=1}^{i-1} (1-\bm{\hat{\alpha_j}}),
\end{equation}

where $\bm{c_i}$ is decoded by color feature $\bm{f_i}$ for each Gaussian points, $\hat{\alpha}$ is the blending weight for each 2D Gaussian point which is the projection of 3D Gaussian points on the image plane.
$N$ is the number of Gaussian points.
Similarly, the depth of pixel $x$ is rendered using alpha blending.

\begin{equation}
    \bm{D}(\bm{x}) = \frac{\sum_{i \in N} d_i 
    \bm{\hat{\alpha_i}} \prod_{j=1}^{i-1} (1-\bm{\hat{\alpha_j}})}{\sum_{i \in N}\bm{\hat{\alpha_i}} \prod_{j=1}^{i-1} (1-\bm{\hat{\alpha_j}})} ,
\end{equation}
where $d_i$ is the distance from the target camera to the Gaussian Point $\bm{P_i}$.
We can also render Radiance Flow between two views of camera $m, n$ using
their camera poses and position of Gaussian points $\bm{\mu_i}$.
%
As mentioned above, we can project pixel $\bm{x} = (u_1, v_1)$ in $m$-th view image
to the $n$-th view by its corresponding depth and their pose transformation:
\begin{equation}
    \bm{D}^n(u_2 , v_2)
    \begin{bmatrix}
     u_2\\
     v_2\\
     1
    \end{bmatrix}
    = 
    \bm{K}\bm{T_{m}^{n}}\bm{K^{-1}}\bm{D}^m(u_1 , v_1)
    \begin{bmatrix}
     u_1\\
     v_1\\
     1
    \end{bmatrix},
    \label{equ:flow_pose}
\end{equation}

where $\bm{T_{m}^{n}}$ is the relative transformation from $m$-th view to $n$-th view, 
$\bm{K}$ is the intrinsic matrix, and $\bm{D^m}(u_1, v_1)$ is the rendered depth of $(u_1, v_1)$ in $m$-th view.

Next, we calculate $\bm{F^{m\rightarrow n}}(\bm{x})$ for pixel $\bm{x} = (u_1, v_1)$ from view $m$ to view $n$:

\begin{equation}
    \bm{F^{m\rightarrow n}}(\bm{x}) = 
    \begin{bmatrix}
     u_2 - u_1\\
     v_2 - v_1
    \end{bmatrix}. 
    \label{equ:flow}
\end{equation}

For other type of Gaussian radiance field such as 2DGS~\citep{huang20242d}, 
we only need to replace the alpha blending based depth $\bm{D}(\bm{x})$ with
corresponding formulation.

\subsection{Flow Distillation Sampling}
\label{sec:method:fds}
Given a collection of images $\{\bm{I^i}\}_{i=1,2,\dots N}$, 
Gaussian Radiance Field typically employs the following loss function for rendering optimization:

\begin{equation}
L = \frac{1}{B}\sum_{i=1}^B(1 - \lambda )L_1 + \lambda L_{D-SSIM} +
\lambda_{normal} L_{n} ,
\end{equation}

where $B$ denotes batch size, 
and $L_n$ represents the normal consistency loss from ~\citep{huang20242d}.
However, when $N$ is small, 
this representation suffers from
overfitting~\citep{li2024dngaussian, 
paliwal2024coherentgs}. 

\begin{wrapfigure}{htbp}{0.5\textwidth} \centering
    \vspace{-2.0em}
    \hspace{2.0em}
    \includegraphics[trim={30pt 0pt 30pt 0pt}, width=0.5\textwidth,height=0.3\textwidth]{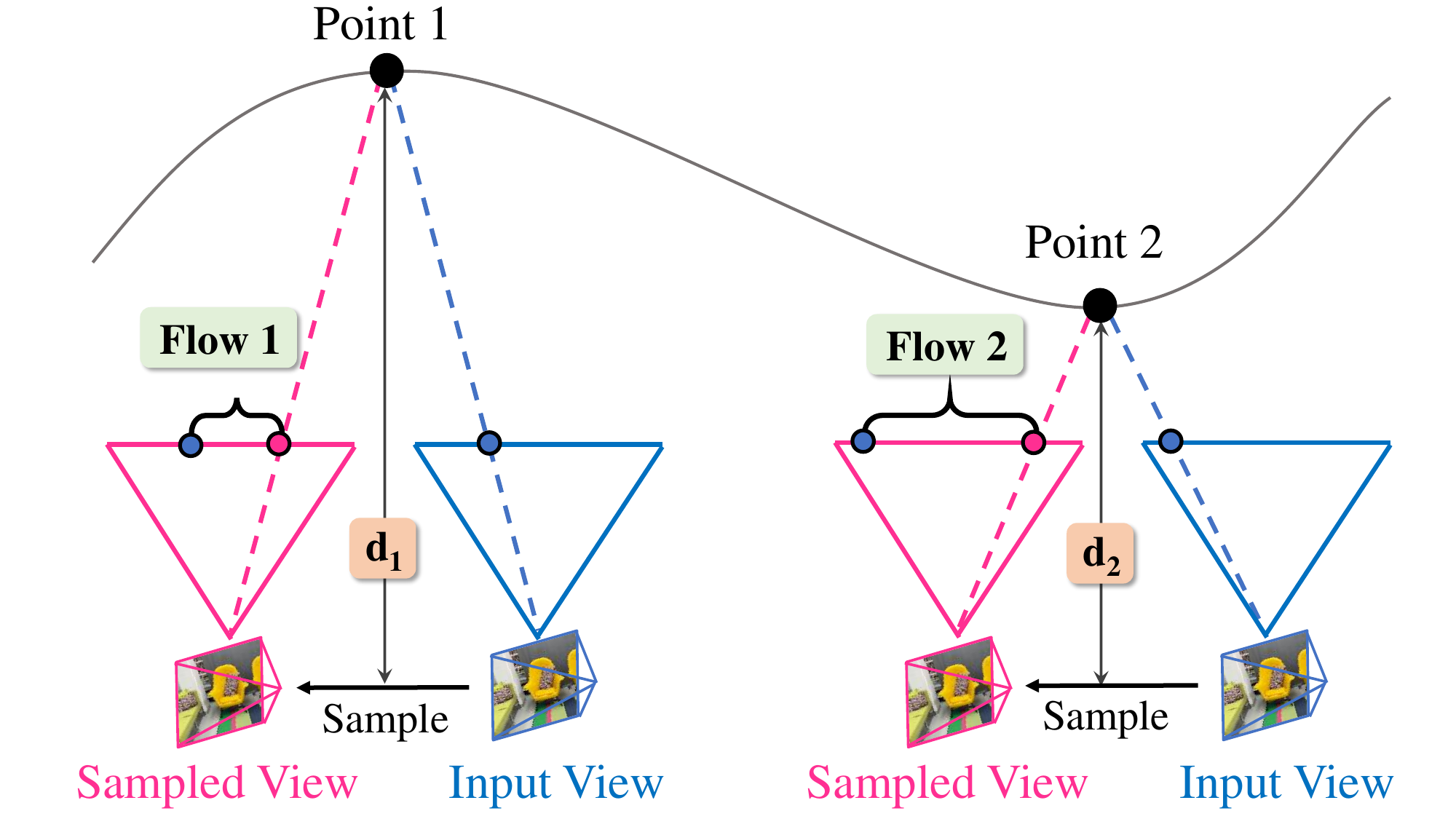}
    \caption{\textbf{Explanation of depth-adaptive translation radius.} A fixed-radius camera sampling scheme may result in significantly different flow values (Flow 1 and Flow 2) in areas with varying depth ($d_1$ and $d_2$).} \label{fig:camera}
    \vspace{-4.0em}
\end{wrapfigure}

We propose the Flow Distillation Sampling (FDS) method to 
incorporate pretrained matching priors into the rendering 
optimization process, thereby mitigating overfitting and 
improving rendering performance.

The camera sampling scheme and loss function design in FDS
are introduced in ~\secref{sec:method:subsubsec:cs} 
and ~\secref{sec:method:subsubsec:fds}.

\subsubsection{Camera Sampling Scheme}
\label{sec:method:subsubsec:cs}

During the training process, FDS randomly
samples unobserved camera views nearby the input view and then incorporates 
the matching priors into these views. 
To utilize the matching prior information between
input view and sampled view more profoundly, 
we propose a camera sampling scheme that ensures sufficient movement to perceive the geometry of the scene while avoiding excessively abrupt motions that could make it difficult for the prior model to match.

Given the world to camera transformation matrix $\bm{T^{i}}$ for the $i$-th input view , 
a small translation disturbance $\bm{t}$ and rotation 
disturbance $\bm{R}$ are applied to get the 
sampled transformation matrix $T^s$, expressed as:
\begin{equation}
\bm{T^s} = ((\bm{T^{i}})^{-1} \bm{E})^{-1}
\label{equ:camera_sampling}
\end{equation}
where $\bm{E}=[\bm{R}, \bm{t}]$ denotes the translation and rotation matrix
in camera coordinates, 
and $t$ is a translation vector satisfying 
$|\bm{t}| = \epsilon_t, \bm{t} = (t_1, t_2, t_3)$,
indicating that unobserved view is sampled along a small
spiral circle with a radius of $\epsilon_t$. 
%
%
As noted in ~\citep{bian2021auto}, the rotation flow in 
image warping is independent of depth. Therefore, we set the 
rotation part to identity matrix. 
So that $E$ is a pure-translation transformation.
For the translation radius $\epsilon_t$, as illustrated in~\figref{fig:camera}, using
a uniform radius for all views leads to varying flows due to
the differences in depths; the closer the depth, the larger the flow. 
To ensure flow consistency across all views 
while maintaining controllable overlap between the input views and sampled views,
we hope to implement a depth-adaptive radius, and
its hyperparameter can be shared across different types of datasets.
which can help to preserve the same flow 
between all input views and their sampled views.

According to ~\eqref{equ:flow_pose}, 
the pure translation transformation ~\citep{bian2021auto} between the input view $i$ and its sampled view $s$ is shown below:

\begin{equation}
    \bm{D^s}(u_2 , v_2)
    \begin{bmatrix}
     u_2\\
     v_2\\
     1
    \end{bmatrix}
    = 
    \bm{D^i}(u_1 , v_1)
    \begin{bmatrix}
     u_1\\
     v_1\\
     1
    \end{bmatrix} 
         +\bm{K}
    \begin{bmatrix}
     t_1\\
     t_2\\
     t_3
    \end{bmatrix} 
    \label{equ:flow_sampling}
\end{equation}

where $K$ is the intrinsic matrix of the camera, after solving the above equation~\citep{bian2021auto}, we get:

\begin{equation}
\begin{bmatrix}
     u_2\\
     v_2\\
    \end{bmatrix} 
= \begin{bmatrix}
\frac{\bm{D_i}(u_1, v_1)u_1+f_xt_1+c_xt_3}{\bm{D_i}(u_1, v_1) + t_3} \\
\frac{\bm{D_i}(u_1, v_1)u_1+f_yt_2+c_xt_3}{\bm{D_i}(u_1, v_1) + t_3} 
\end{bmatrix} 
\end{equation}

We set $ t_3 = 0 $ in our camera sampling scheme and assume camera intrinsic parameters: $f_x\approx f_y=f$. 
The radiance flow $\bm{F^{i\rightarrow s}}(u_1, v_1) = \begin{bmatrix}
     u_2 - u_1\\
     v_2 - v_1\\
    \end{bmatrix} $from the input view $i$ to its sampled view $s$ is shown below:

\begin{equation}
\bm{F^{i\rightarrow s}}(u_1, v_1) = 
    \begin{bmatrix}
    \frac{f}{\bm{D_i}(u_1, v_1)}t_1 \\
    \frac{f}{\bm{D_i}(u_1, v_1)}t_2
    \end{bmatrix} 
\label{equ:flow_sampling_2}
\end{equation}

We aim to keep the value of
$||\bm{F^{i\rightarrow s}}(u_1, v_1)||_2$  constant for the pixel $x = (u_1, v_1)$
during each camera sampling. By setting $||\bm{F^{i\rightarrow s}}(u_1, v_1)||_2 = \sigma $ and
incorporating this with ~\eqref{equ:flow_sampling_2}, 
we get:

\begin{equation}
    \epsilon_t = \sqrt{t_1^2 + t_2^2}  =  \sigma  \frac{\bm{D_i}(u_1, v_1)}{f}
\end{equation}

Thus, the radius of translation in our camera sampling 
is defined as 
$\epsilon_t = \sigma \frac{\bm{D_i}(u_1, v_1)}{f} $ 
which helps maintain stable flow.
The parameter $\sigma$ can be tuned as a hyperparameter,
which represents the mean radiance flow between the input view 
and its sampled view.
Given that pixel depths vary within an image, 
we use the mean depth $ \bar{\bm{D_i}} $ of the image 
and set the radius of our translation
$\epsilon_t = \sigma  \frac{\bar{\bm{D_i}}}{f} $.

\begin{equation}
    \bm{t} = \begin{bmatrix}
        \sigma  \frac{\bar{\bm{D_i}}}{f} sin(2 \pi \xi ) \\
        \sigma \frac{\bar{\bm{D_i}}}{f} cos(2 \pi \xi ) \\
        0
    \end{bmatrix}
\end{equation}

Where $\xi ~\sim  U(0, 1) $ is a uniform 
random value between $[0, 1]$ during training.


\subsubsection{Flow Distillation Sampling Loss}
\label{sec:method:subsubsec:fds}
Specifically, to optimize the rendered results of the $i$-th view, 
every iteration, we sample an unobserved view $s$ near the $i$-th input training camera 
to render the color image $\bm{C^s}$. The Radiance Flow $\bm{F^{i\rightarrow s}(p)}$,
guided by \equref{equ:flow}, 
benefits from the introduced matching priors. 
We define the Prior Flow $\bm{\overline{F}^{i\rightarrow s}}$ as:

\begin{equation}
\bm{\overline{F}^{i\rightarrow s}} = \mathcal{M}_{\theta}(\bm{I^i}, \bm{C^s}),  
\label{equ:mflow}
\end{equation}

where 
$\theta$ represents the parameters of pretrained network $\mathcal{M}$,
$\bm{I^i}$ is the rgb ground truth of view $i$, and $\bm{C^s}$ is the 
rendered image of sampled view $s$. 
Instead of using $\bm{C^i}$ to generate Prior Flow, we find that using 
$\bm{I^i}$ can 
help us to remove the floaters in $\bm{C^s}$ as shown in ~\tabref{tab:ablation_fds}.
However, both $\bm{F^{i\rightarrow s}}$ and 
$\bm{\overline{F}^{i\rightarrow s}}$ face challenges:
\begin{itemize}
    \item $\bm{F^{i\rightarrow s}}$, 
    derived from \equref{equ:flow}, 
    is inaccurate due to the incorrect positioning 
    of Gaussian points.
    \item $\bm{\overline{F}^{i\rightarrow s}}$, derived from \equref{equ:mflow}, 
    suffers from the blurred rendering quality of $\bm{C^s}$.
\end{itemize}

Despite these inaccuracies, we observe 
that $\bm{\overline{F}^{i\rightarrow s}}$ is
more robust and precise during training compared with $\bm{F^{i\rightarrow s}}$.
Motivated by this observation, we aim to utilize $\bm{\overline{F}^{i\rightarrow s}}$
to refine $\bm{F^{i\rightarrow s}}$, which can make $\bm{F^{i\rightarrow s}}$ more accurate, 
enhancing the positioning of Gaussian points and subsequently 
improving the rendering quality of $\bm{C^s}$. 
This process  also leads to a more accurate $\bm{\overline{F}^{i\rightarrow s}}$. 
Based on these observations, we propose the FDS loss, which distills
matching priors from a pretrained deep model through mutual 
refinement of the two flows:

\begin{equation}
L_{fds} =  || \bm{\overline{F}^{i\rightarrow s}} - \bm{F^{i\rightarrow s}} ||_2
\label{equ:fds}
\end{equation}

\begin{algorithm}[h]
  \caption{Flow Distillation Sampling
   \label{algo:fds}} %

  \begin{algorithmic}[1]
\Require
      A batch of input training image: $\{\bm{I_i}\}_{i=1}^{N}$,  
     Transformation Matrix: $\{\bm{T_i}\}_{i=1}^{N}$, 
     Prior Matching Network $\mathcal{M}_{\theta}$, 
     Gaussian Points $\{\bm{P}_{i}\}_{i=1}^M$ with $\{\bm{r_i}, \bm{t_i}, \bm{f_i}, \bm{\mu_i}, \bm{\alpha_i}\}$.
\Ensure
      $L_{fds}$

 \For {$i$ in $\{1, 2, \dots, B\}$}
 \State \textcolor{blue}{$\xi ~\sim  U(0, 1), \bm{R}\leftarrow \bm{I}$}
 \State  \textcolor{blue}{$t_1 \leftarrow  \sigma \frac{\bar{\bm{D_i}}}{f} sin(2 \pi \xi ), 
 t_2 \leftarrow  \sigma  \frac{\bar{\bm{D_i}}}{f} cos(2 \pi \xi ), t_3 \leftarrow 0$}
 \State  \textcolor{blue}{$\bm{E}  \leftarrow [\bm{R}, \bm{t}], \bm{T^s} \leftarrow ((\bm{T^{i})}^{-1} \bm{E})^{-1}$}
 \State $\bm{C^s}, \bm{D^s} \leftarrow Render(\bm{T^s}, \bm{P})$
 \State $\bm{C^i}, \bm{D^i} \leftarrow Render(\bm{T^i}, \bm{P})$
 \State \textcolor{olive}{$\bm{X^s} \leftarrow \bm{KT_{i}^{s}K}^{-1}\bm{D^i}(X^i)X^i/ \bm{D}(\bm{X^s})$}
 \State \textcolor{olive}{$\bm{F^{i\rightarrow s}} \leftarrow  \bm{X^s} - \bm{X^i} $}
 \State  \textcolor{olive}{$\bm{\overline{F}^{i\rightarrow s}} \leftarrow  \mathcal{M}_{\theta}(\bm{I^i}, \bm{C^s})$}
 \State  \textcolor{olive}{$L_{fds} \leftarrow L_{fds} + 1/B || \bm{\overline{F}^{i\rightarrow s}} 
    - \bm{F^{i\rightarrow s}}  ||_2$}
 
 \EndFor
 \State \Return $L_{fds}$
  \end{algorithmic}

\end{algorithm}

To maintain training stability and reduce computational complexity, 
we detach $\bm{\overline{F}^{i\rightarrow s}}$ from $\bm{P_i}$ 
when calculating loss. 
This prevents the propagation of gradients, as computing them is 
resource-intensive and time-consuming, as noted in \citep{poole2022dreamfusion}.
Additionally, this helps to ensure that $\bm{\overline{F}^{i\rightarrow s}}$ is not directly influenced by the less reliable $\bm{F^{i\rightarrow s}}$.

In summary, the procedures of our proposed 
Flow Distillation Sampling are presented 
in Algorithm~\ref{algo:fds}. The overall training loss is:

\begin{equation}
    L = \frac{1}{B}\sum_{i=1}^B(1 - \lambda )L_1 + \lambda L_{D-SSIM} + \lambda_{normal} L_{n} + \lambda_{fds} L_{fds}.
\end{equation}

\section{Experiments}

\subsection{Setups}
\subsubsection{Implementation Details}
We apply our FDS method to two types of 3DGS: 
the original 3DGS, and 2DGS~\citep{huang20242d}. 
The number of iterations in our optimization 
process is 35,000.
We follow the default training configuration 
and apply our FDS method after 15,000 iterations,
then we add normal consistency loss for both
3DGS and 2DGS after 25000 iterations.
The weight for FDS, $\lambda_{fds}$, is set to 0.015,
the $\sigma$ is set to 23,
and the weight for normal consistency is set to 0.15
for all experiments. 
We removed the depth distortion loss in 2DGS 
because we found that it degrades its results in indoor scenes.
The Gaussian point cloud is initialized using Colmap
for all datasets.
We tested the impact of 
using Sea Raft~\citep{wang2025sea} and 
Raft\citep{teed2020raft} on FDS performance.
Due to the blurriness of the ScanNet dataset, 
additional prior constraints are required.
Thus, we incorporate normal prior supervision 
on the rendered normals 
in ScanNet (V2) dataset by default.
The normal prior is predicted by the Stable Normal 
model~\citep{ye2024stablenormal}
across all types of 3DGS.
The entire framework is implemented in 
PyTorch~\citep{paszke2019pytorch}, 
and all experiments are conducted on 
a single NVIDIA 4090D GPU.

\begin{figure}[t] \centering
    \makebox[0.16\textwidth]{\scriptsize Input}
    \makebox[0.16\textwidth]{\scriptsize 3DGS}
    \makebox[0.16\textwidth]{\scriptsize 2DGS}
    \makebox[0.16\textwidth]{\scriptsize 3DGS + FDS}
    \makebox[0.16\textwidth]{\scriptsize 2DGS + FDS}
    \makebox[0.16\textwidth]{\scriptsize GT (Depth)}

    \includegraphics[width=0.16\textwidth]{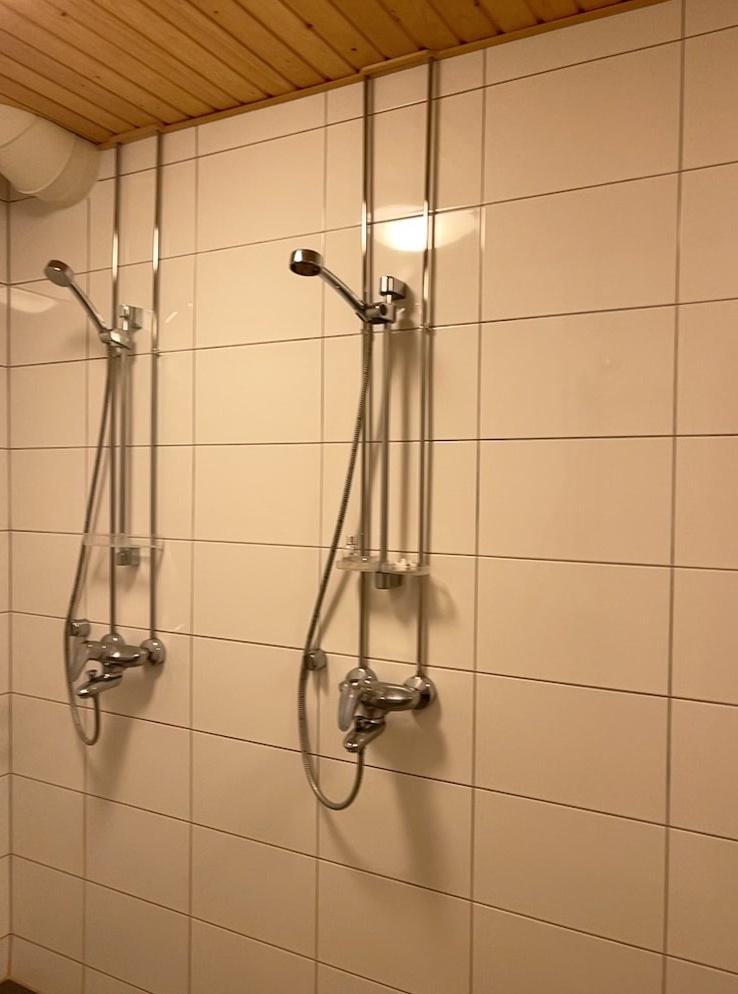}
    \includegraphics[width=0.16\textwidth]{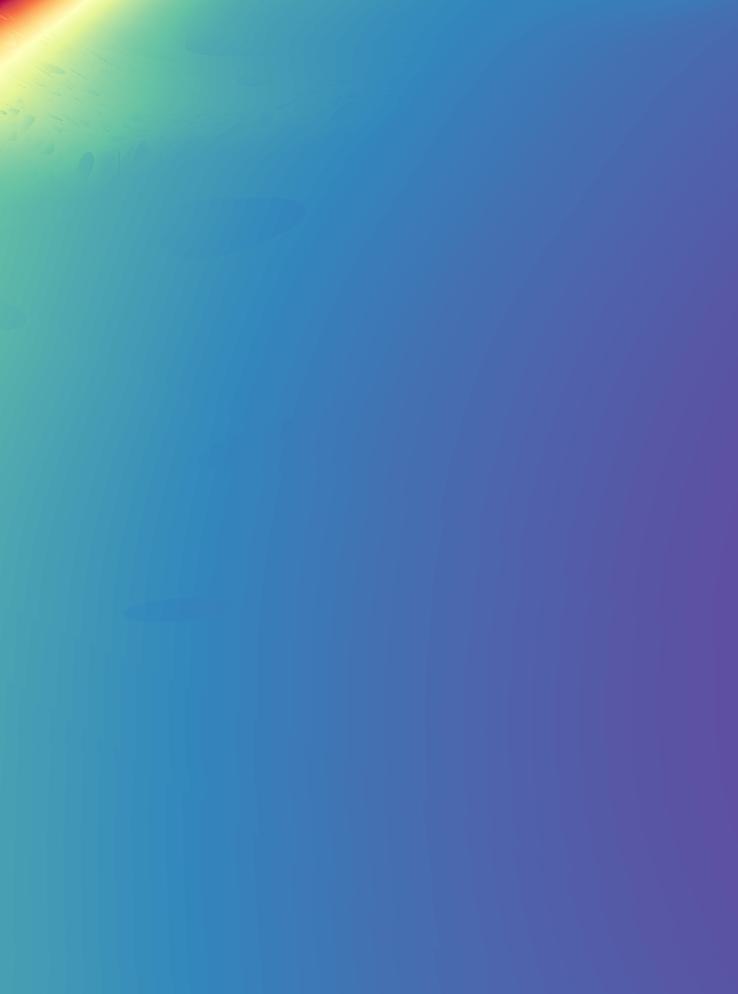}
    \includegraphics[width=0.16\textwidth]{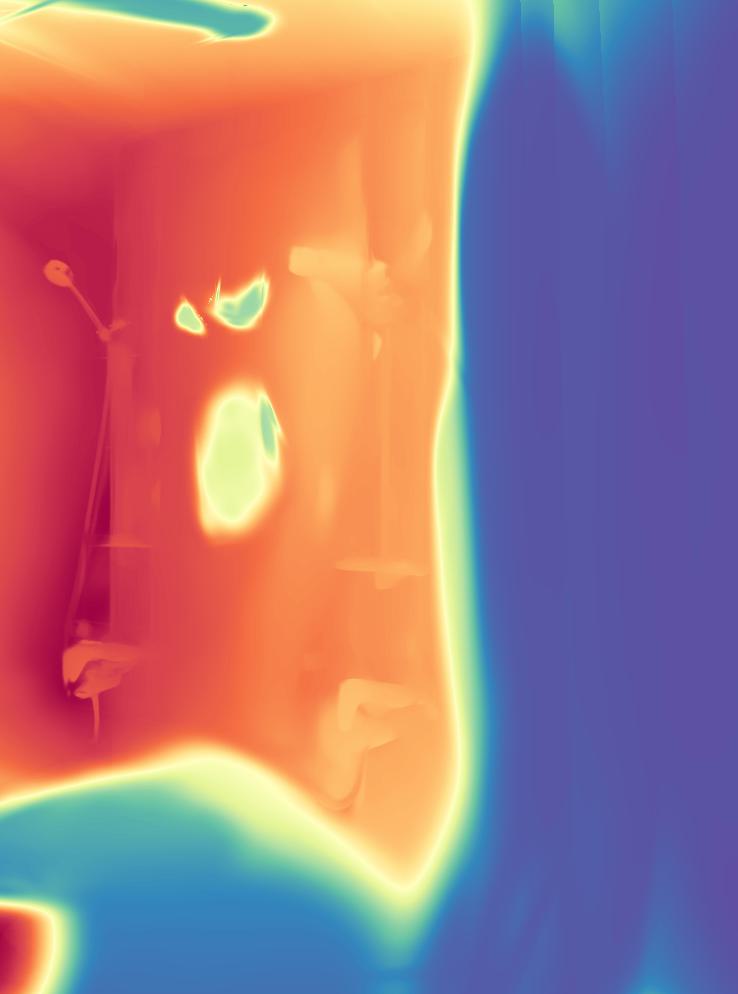}
    \includegraphics[width=0.16\textwidth]{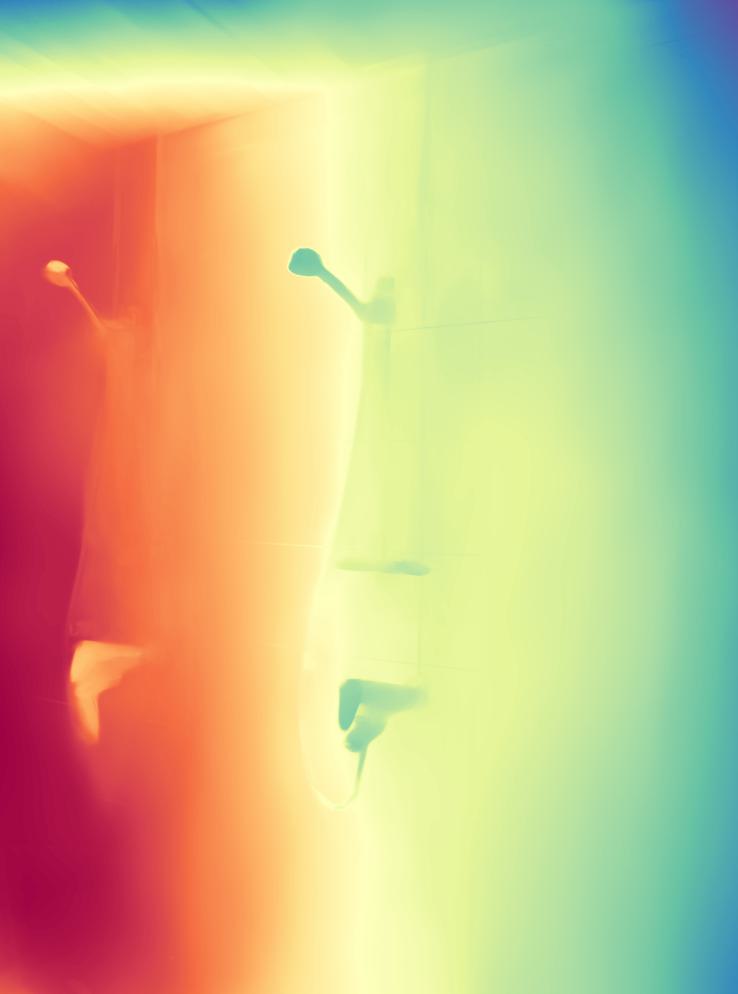}
    \includegraphics[width=0.16\textwidth]{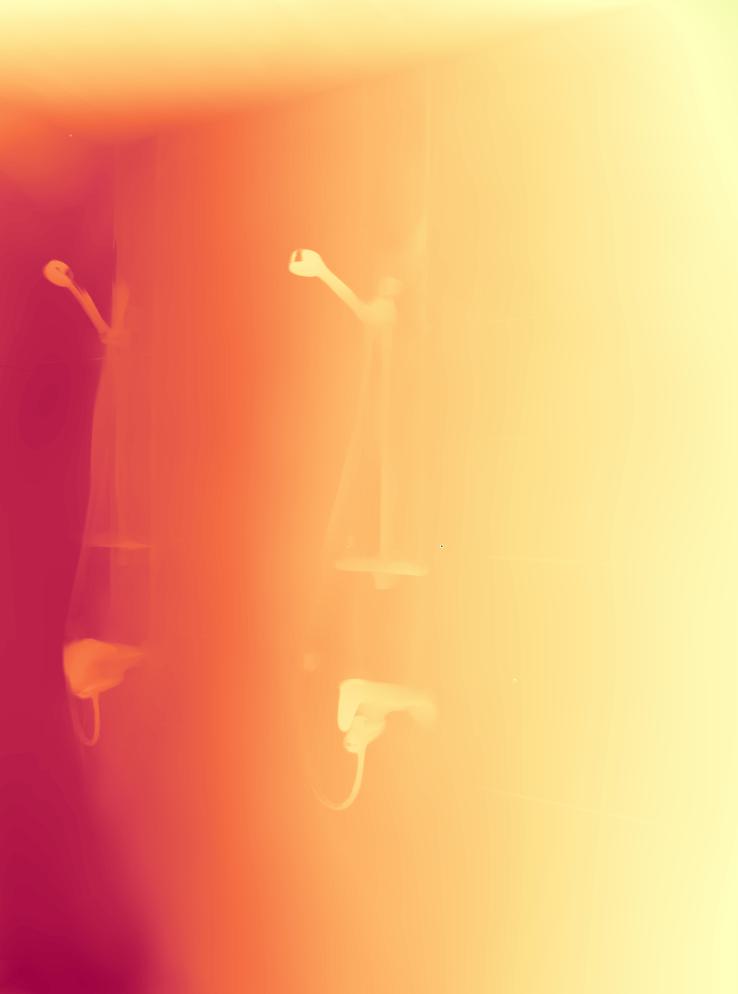}
    \includegraphics[width=0.16\textwidth]{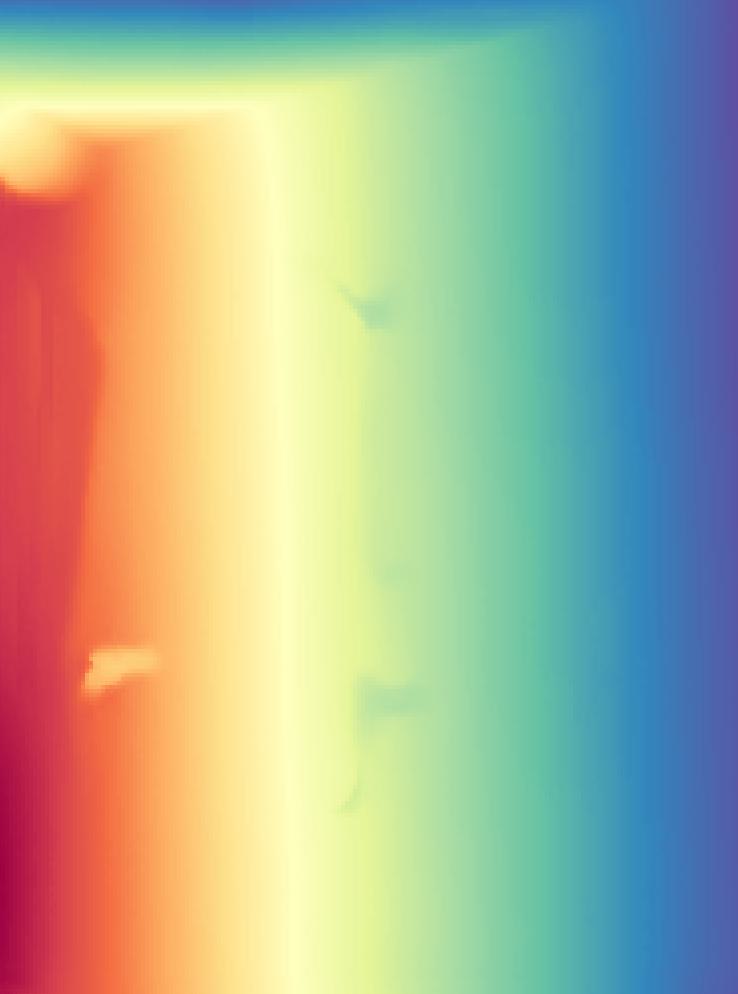} \\

     \includegraphics[width=0.16\textwidth]{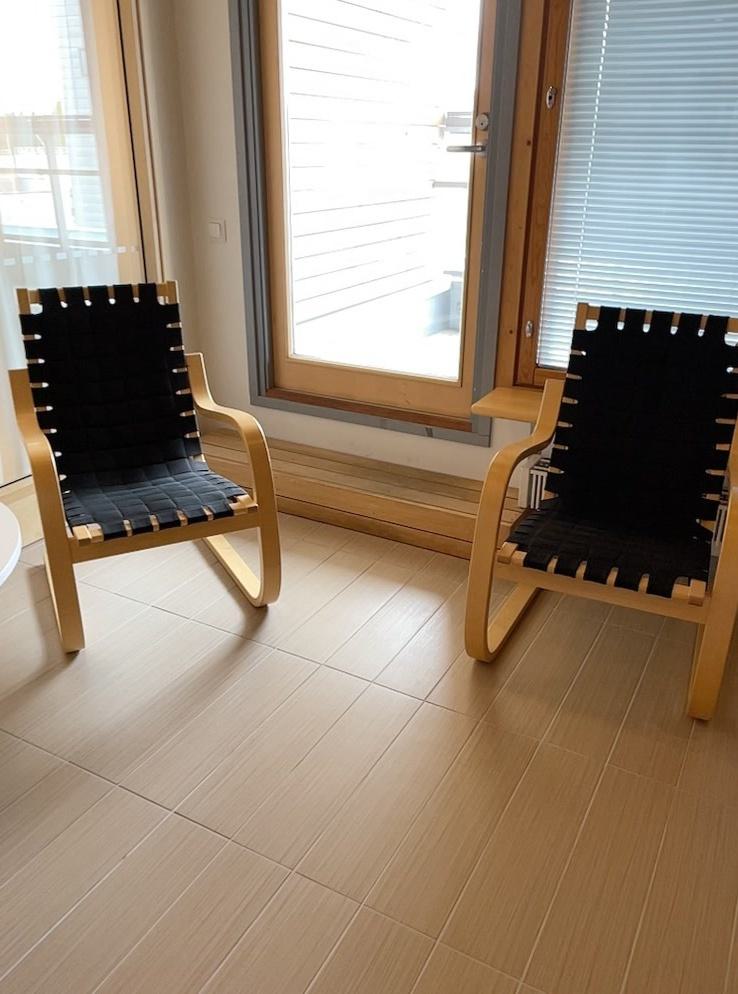}
    \includegraphics[width=0.16\textwidth]{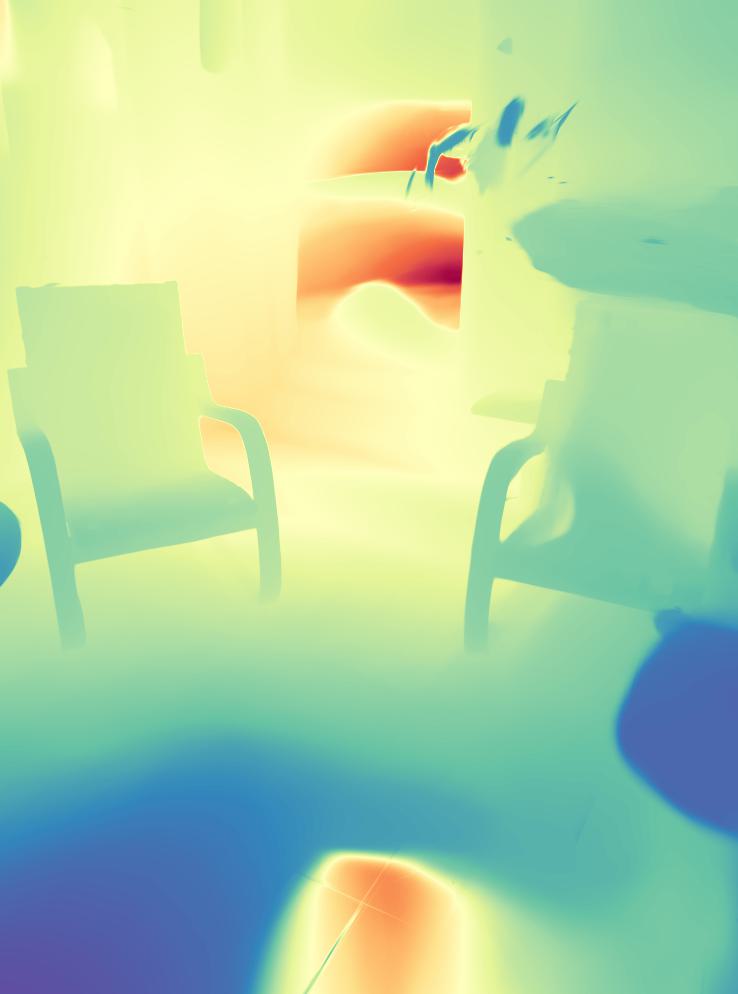}
    \includegraphics[width=0.16\textwidth]{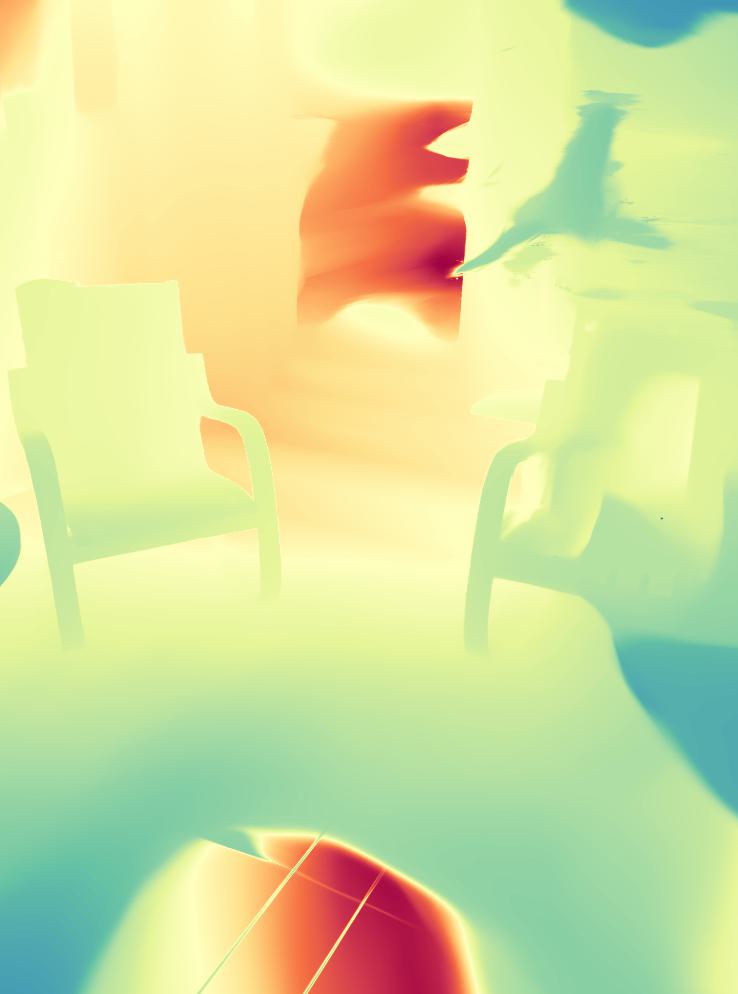}
    \includegraphics[width=0.16\textwidth]{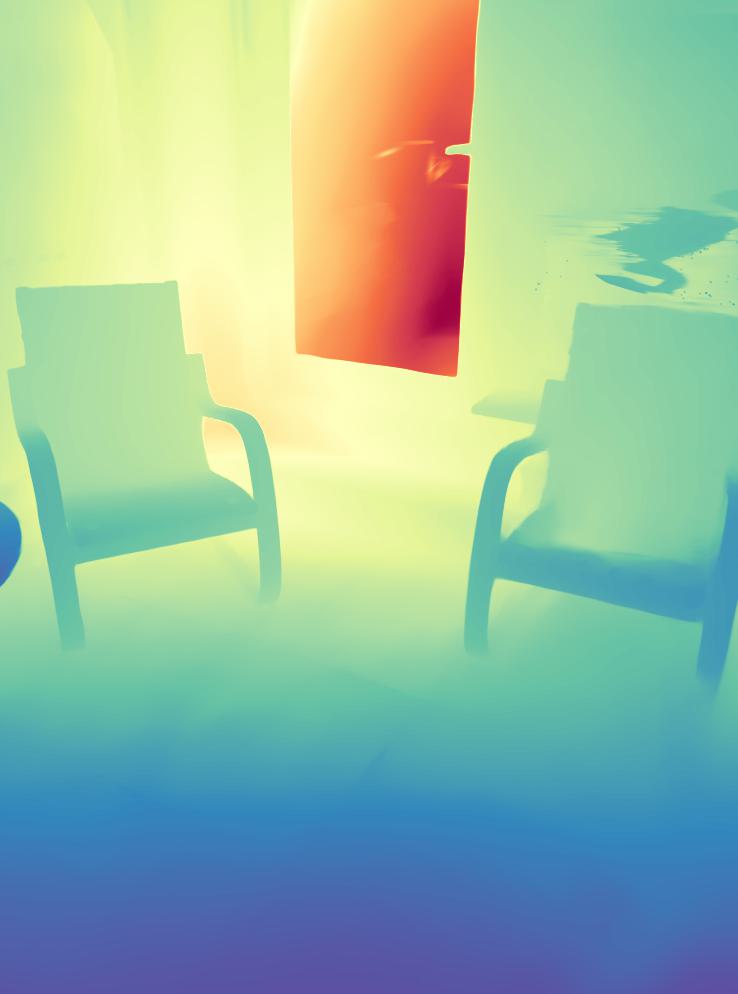}
    \includegraphics[width=0.16\textwidth]{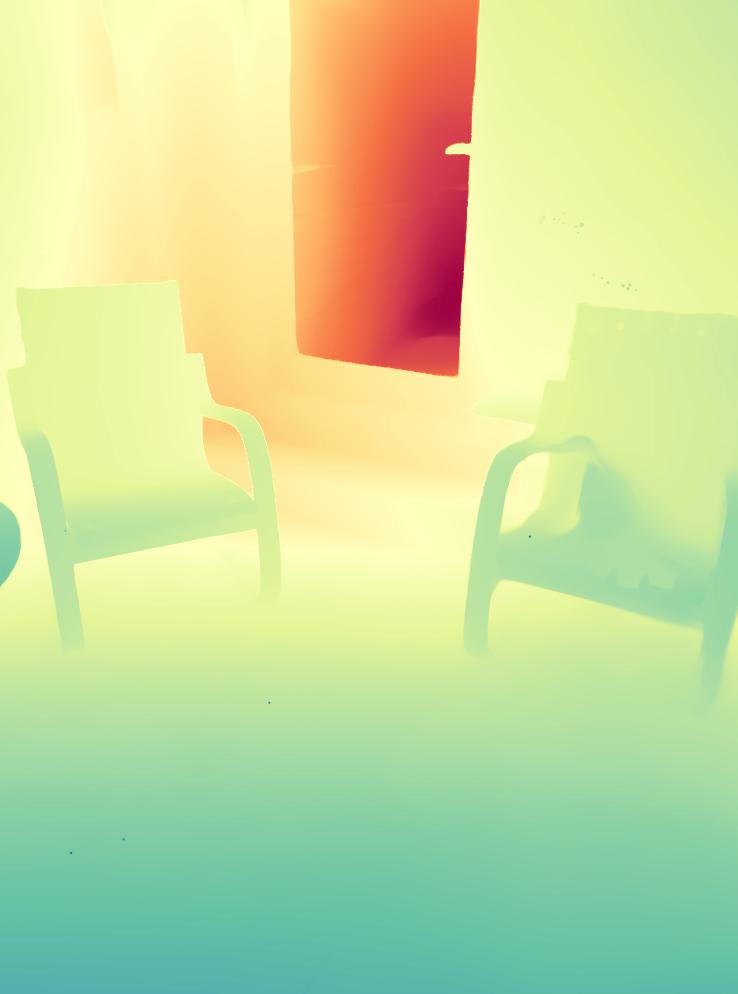}
    \includegraphics[width=0.16\textwidth]{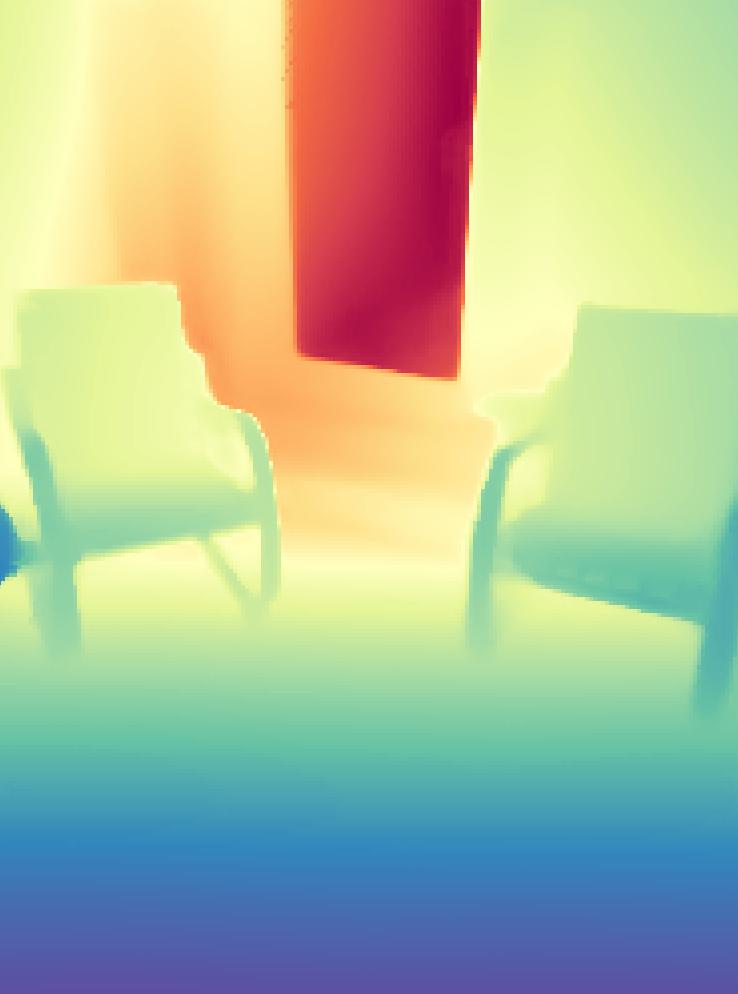} \\

    \includegraphics[width=0.16\textwidth]{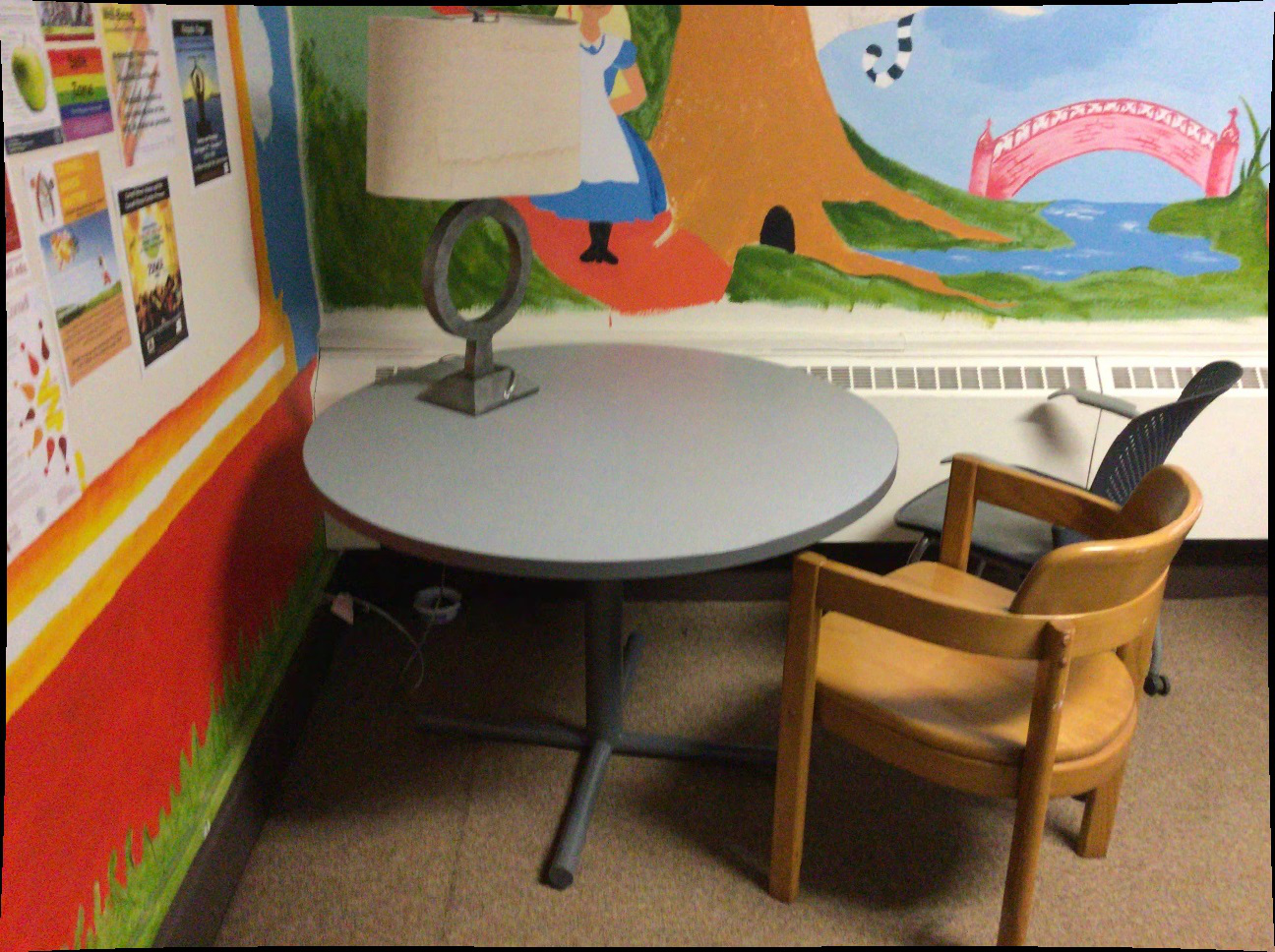}
    \includegraphics[width=0.16\textwidth]{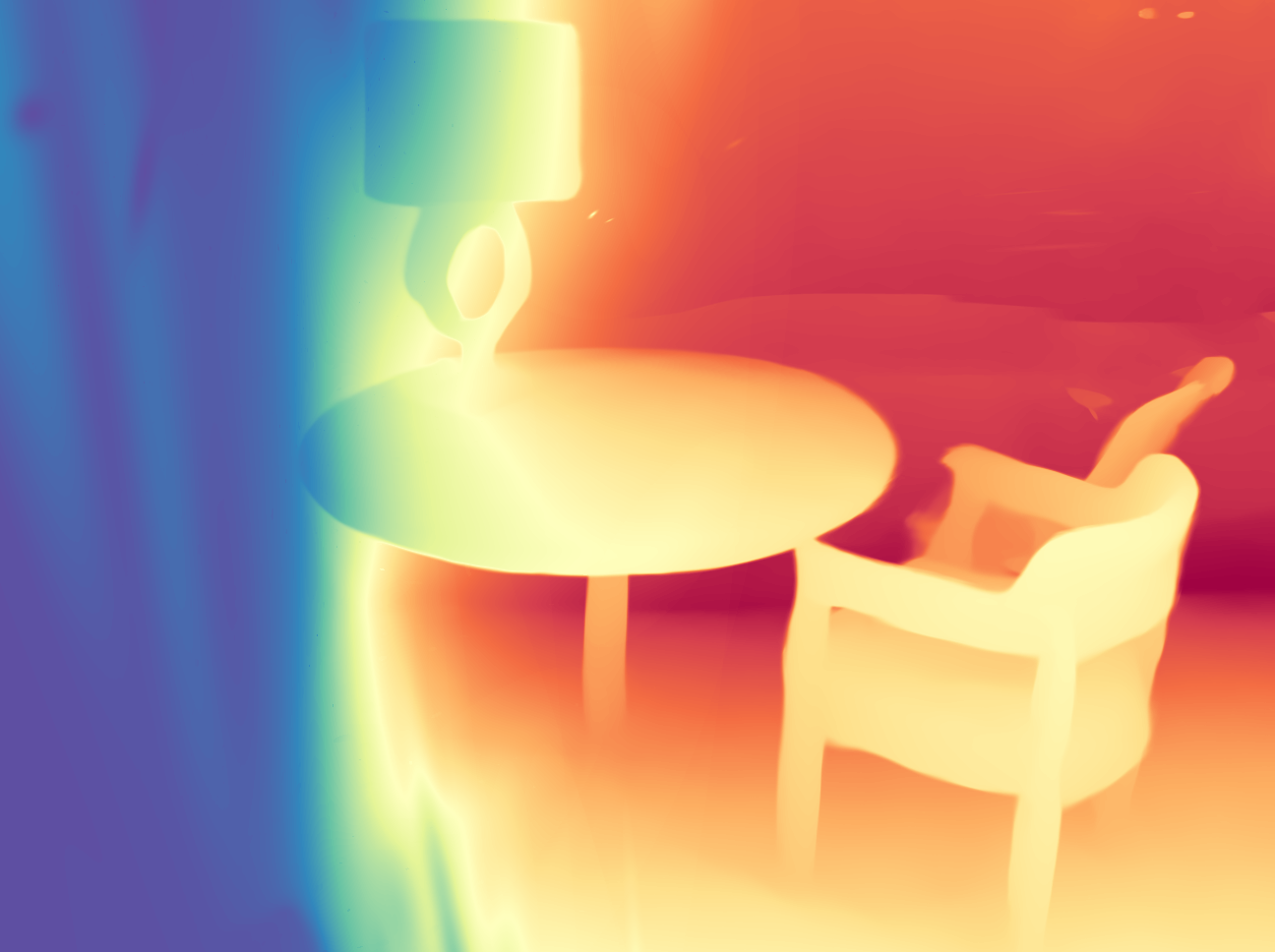}
    \includegraphics[width=0.16\textwidth]{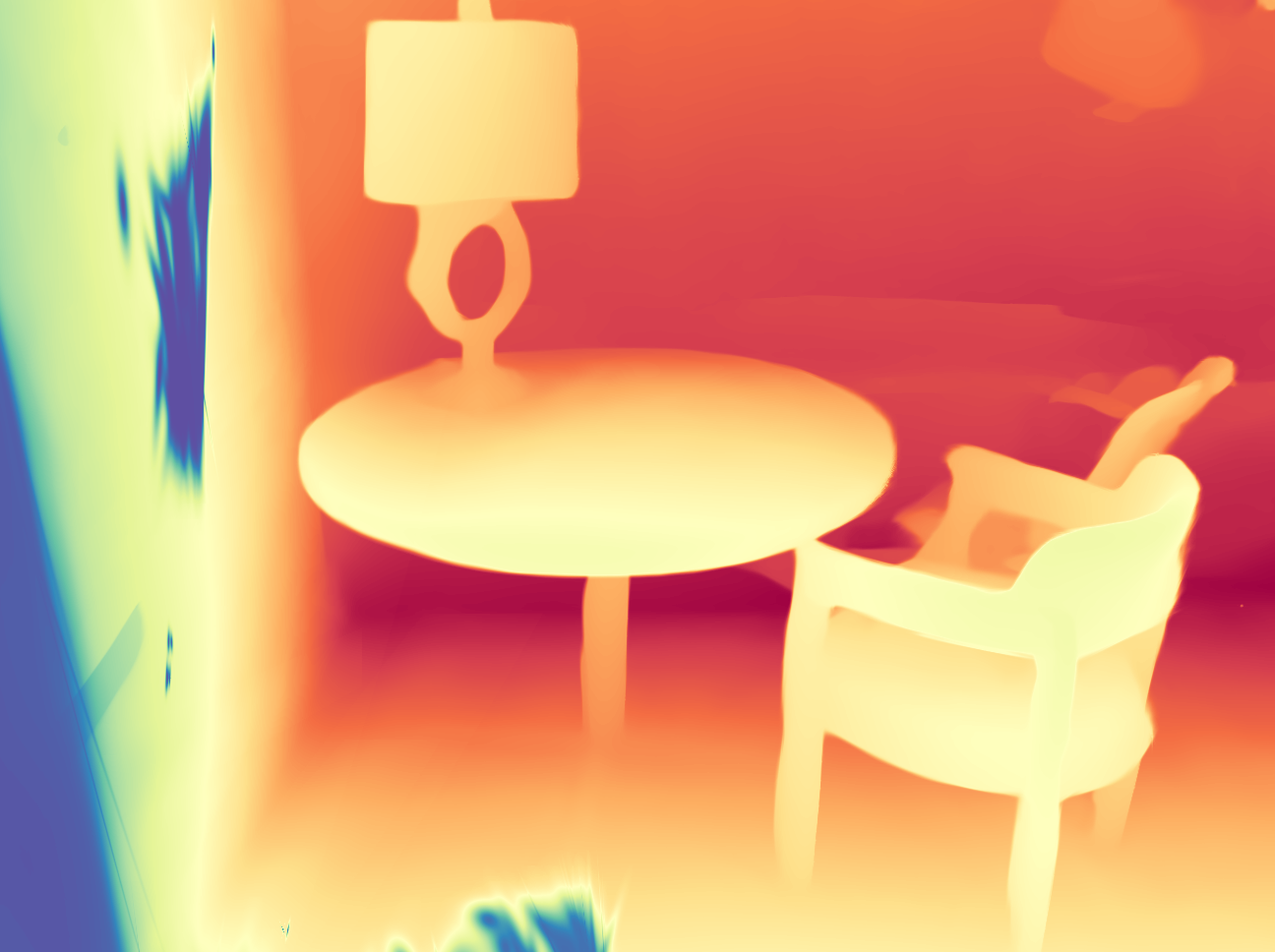}
    \includegraphics[width=0.16\textwidth]{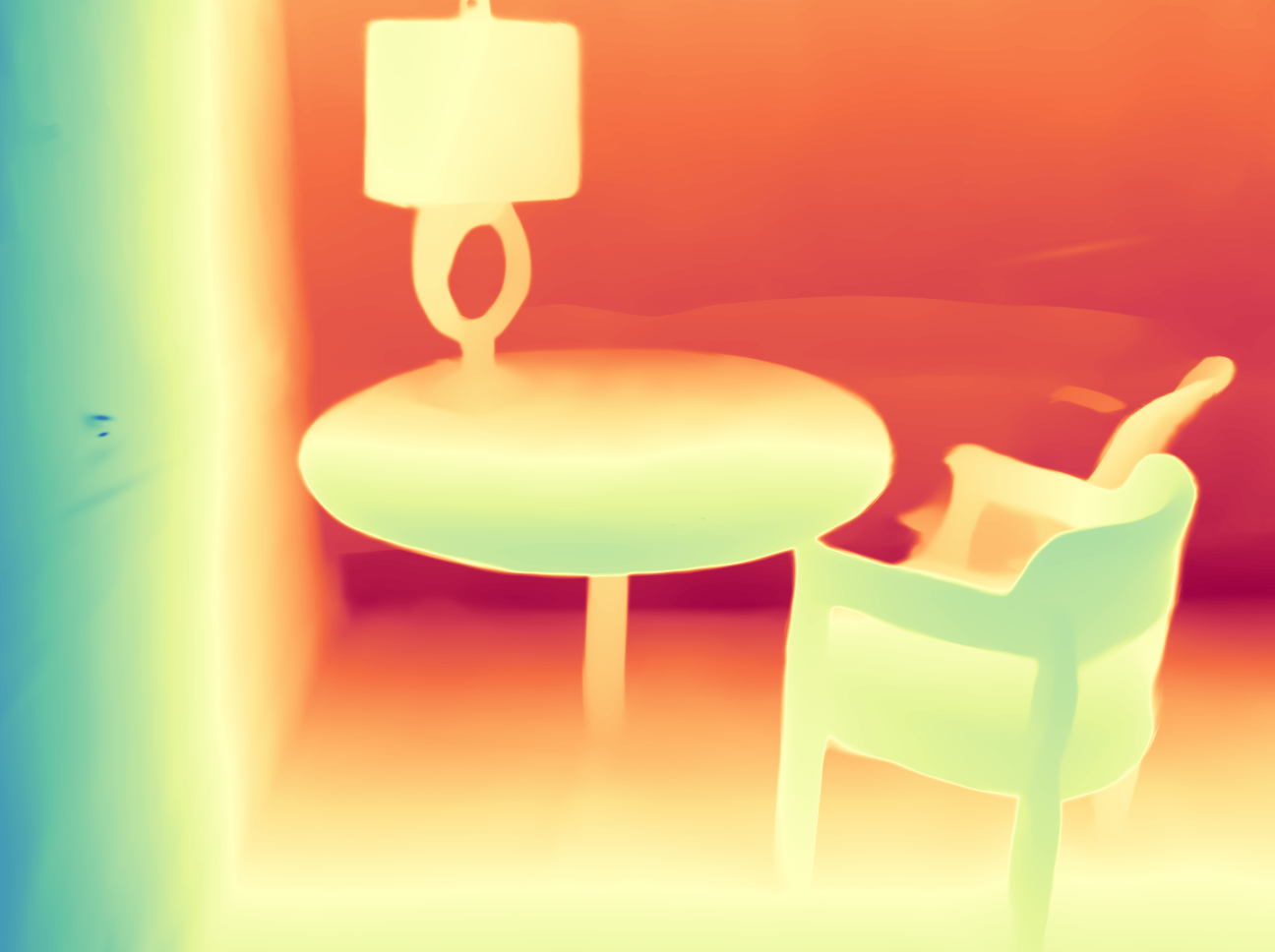}
    \includegraphics[width=0.16\textwidth]{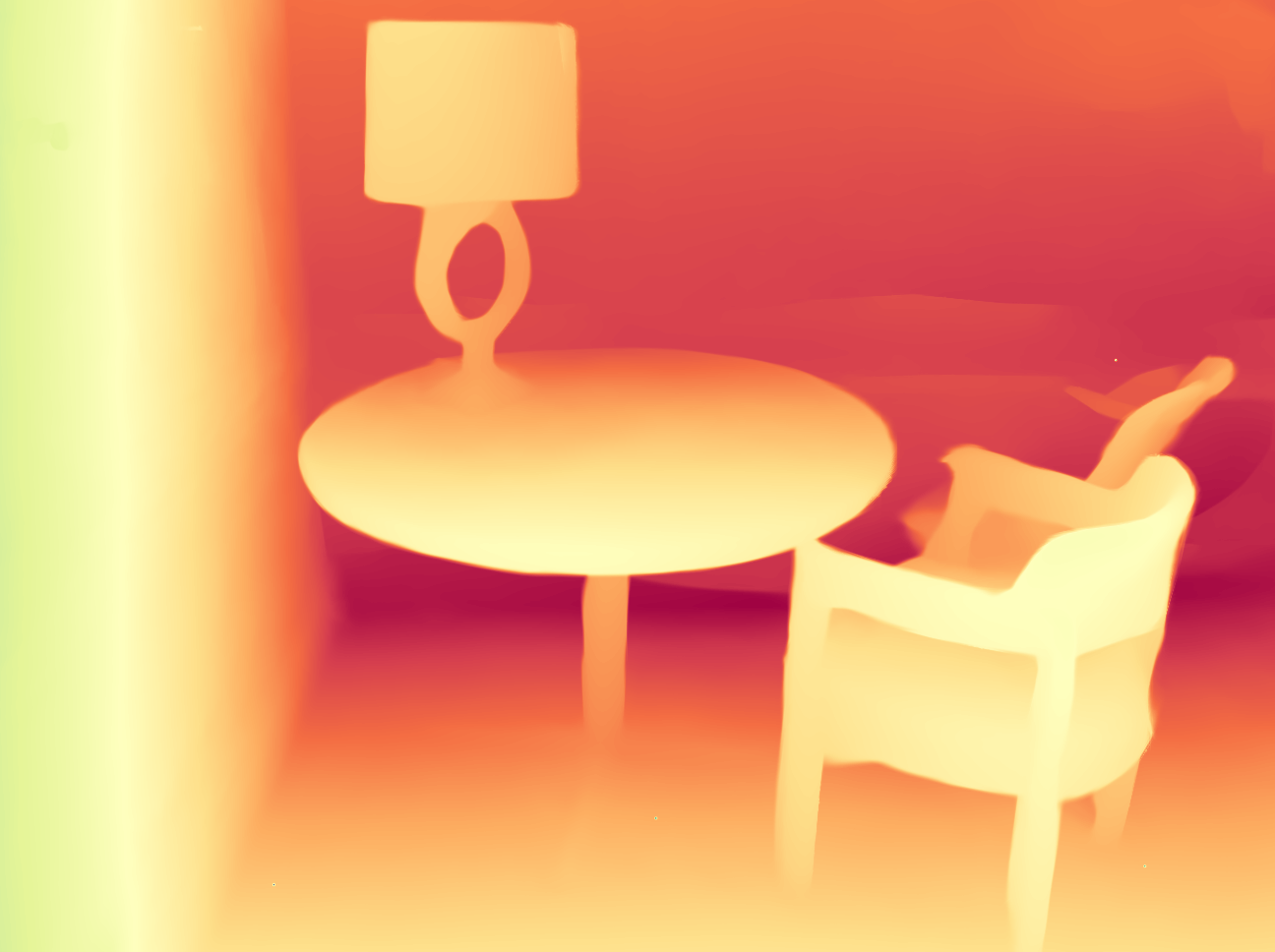}
    \includegraphics[width=0.16\textwidth]{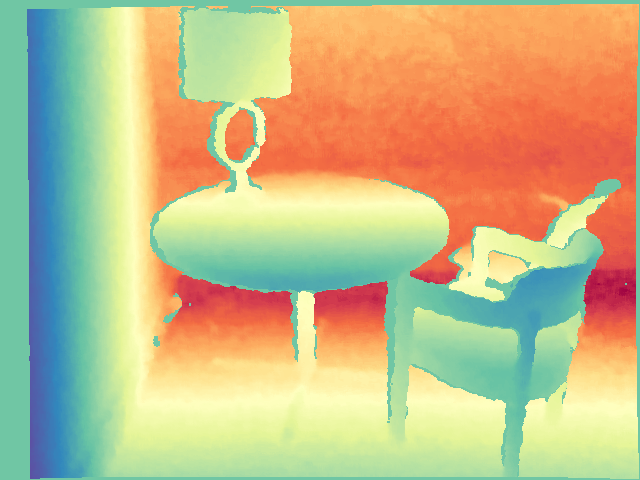} \\

    \includegraphics[width=0.16\textwidth]{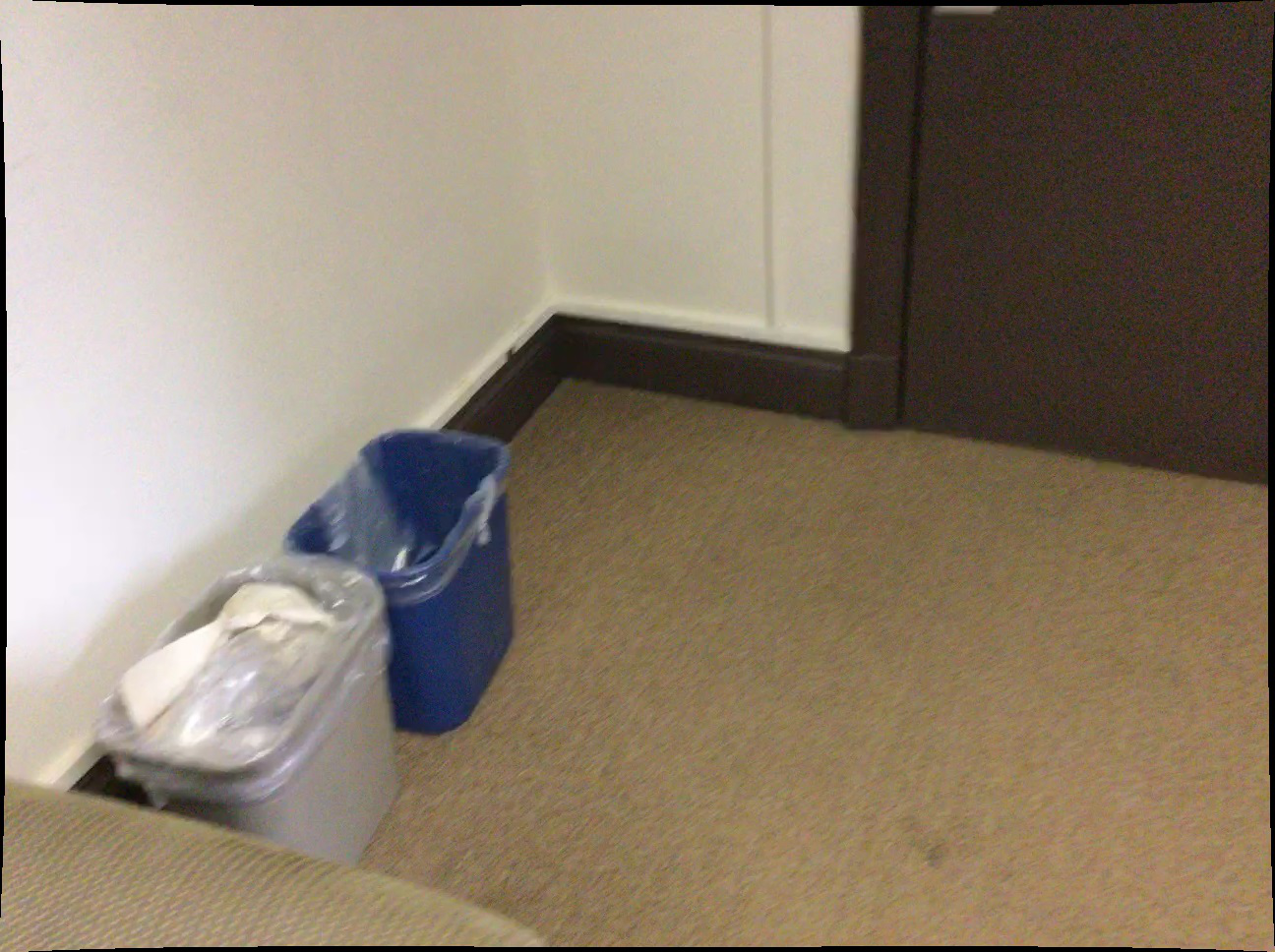}
    \includegraphics[width=0.16\textwidth]{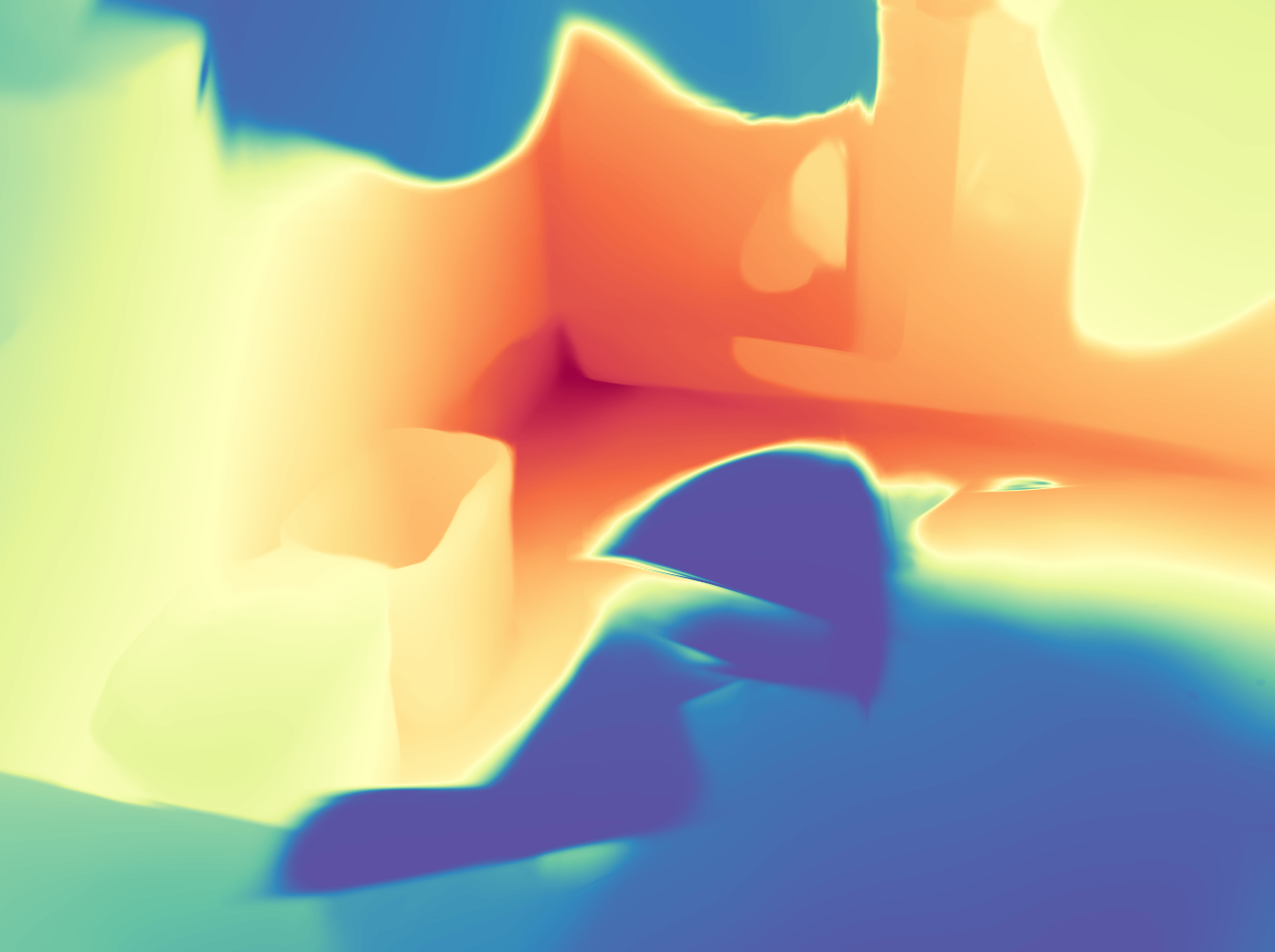}
    \includegraphics[width=0.16\textwidth]{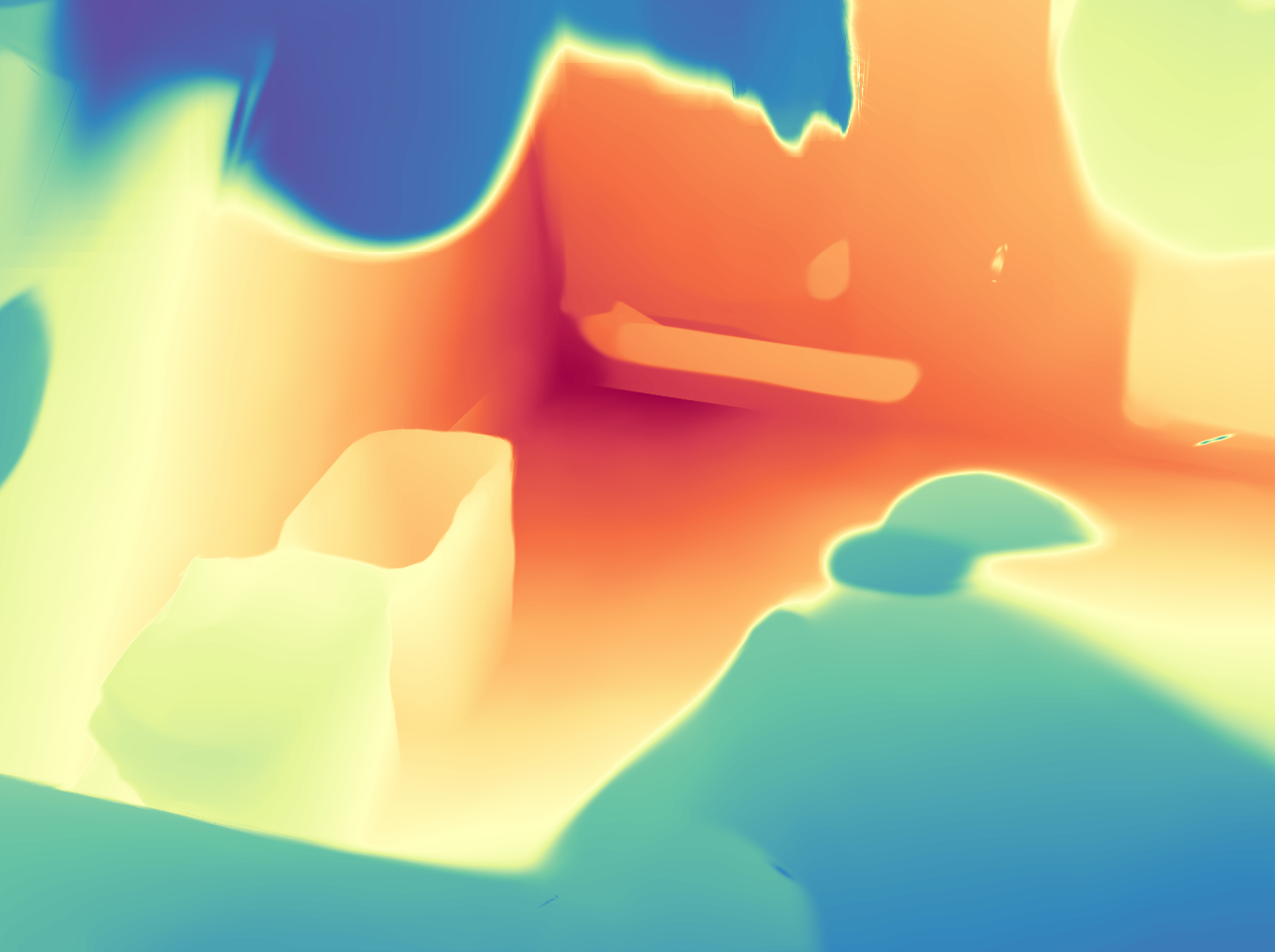}
    \includegraphics[width=0.16\textwidth]{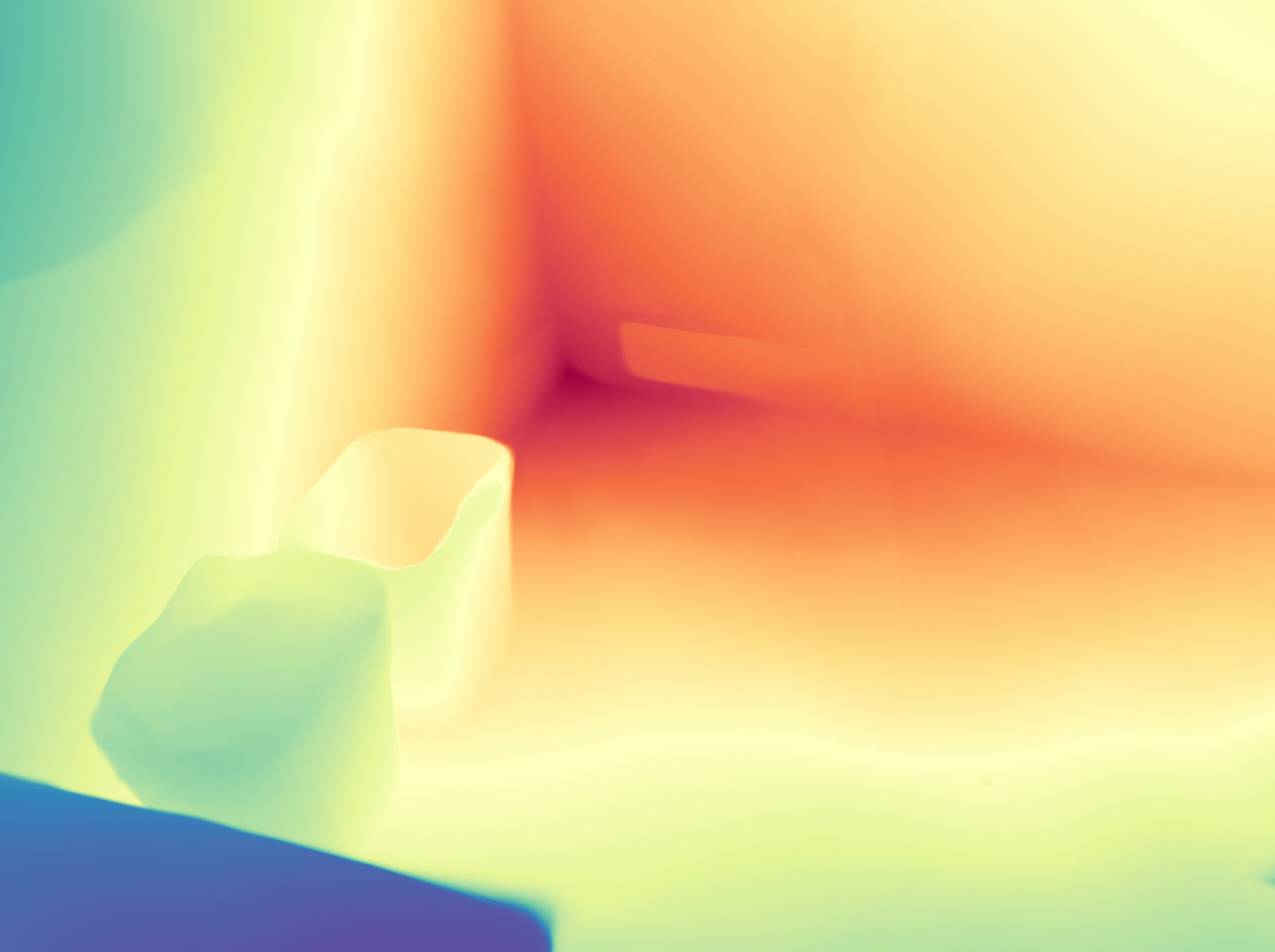}
    \includegraphics[width=0.16\textwidth]{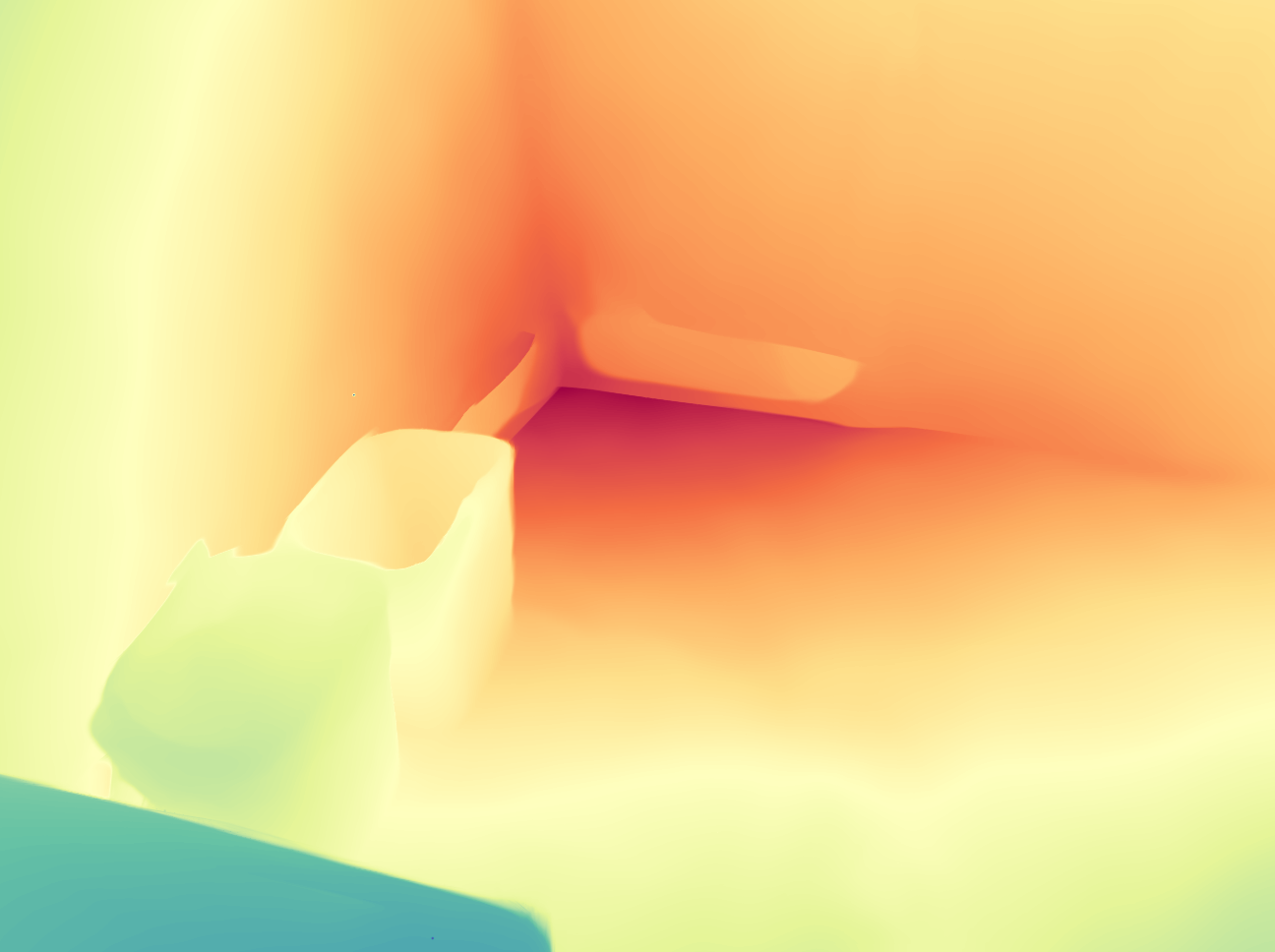}
    \includegraphics[width=0.16\textwidth]{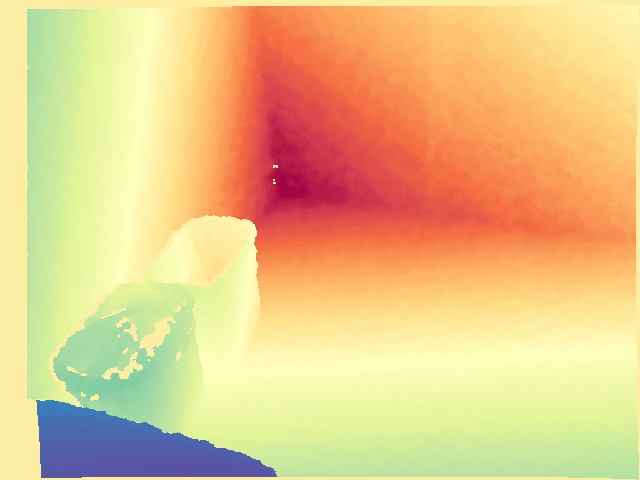} \\

    \caption{\textbf{Comparison of depth reconstruction on Mushroom and ScanNet datasets.} The original
    3DGS or 2DGS model equipped with FDS can remove unwanted floaters and reconstruct
    geometry more preciously.}
    \label{fig:compare}
\end{figure}

\subsubsection{Datasets and Metrics}

We evaluate our method for 3D reconstruction 
and novel view synthesis tasks on
\textbf{Mushroom}~\citep{ren2024mushroom},
\textbf{ScanNet (v2)}~\citep{dai2017scannet}, and 
\textbf{Replica}~\citep{replica19arxiv}
datasets,
which feature challenging indoor scenes with both 
sparse and dense image sampling.
The Mushroom dataset is an indoor dataset 
with sparse image sampling and two distinct 
camera trajectories. 
We train our model on the training split of 
the long capture sequence and evaluate 
novel view synthesis on the test split 
of the long capture sequences.
Five scenes are selected to evaluate our FDS, 
including "coffee room", "honka", "kokko", 
"sauna", and "vr room". 
ScanNet(V2)~\citep{dai2017scannet}  consists of 1,613 indoor scenes
with annotated camera poses and depth maps. 
We select 5 scenes from the ScanNet (V2) dataset, 
uniformly sampling one-tenth of the views,
following the approach in ~\citep{guo2022manhattan}.
To further improve the geometry rendering quality of 3DGS, 
Replica~\citep{replica19arxiv} contains small-scale 
real-world indoor scans. 
We evaluate our FDS on five scenes from 
Replica: office0, office1, office2, office3 and office4,
selecting one-tenth of the views for training.
The results for Replica are provided in the 
supplementary materials.
To evaluate the rendering quality and geometry 
of 3DGS, we report PSNR, SSIM, and LPIPS for 
rendering quality, along with Absolute Relative Distance 
(Abs Rel) as a depth quality metrics.
Additionally, for mesh evaluation, 
we use metrics including Accuracy, Completion, 
Chamfer-L1 distance, Normal Consistency, 
and F-scores.

\subsection{Results}
\subsubsection{Depth rendering and novel view synthesis}
The comparison results on Mushroom and 
ScanNet are presented in \tabref{tab:mushroom} 
and \tabref{tab:scannet}, respectively. 
Due to the sparsity of sampling 
in the Mushroom dataset,
challenges are posed for both GOF~\citep{yu2024gaussian} 
and PGSR~\citep{chen2024pgsr}, 
leading to their relative poor performance 
on the Mushroom dataset.
Our approach achieves the best performance 
with the FDS method applied during the training process.
The FDS significantly enhances the 
geometric quality of 3DGS on the Mushroom dataset, 
improving the "abs rel" metric by more than 50\%.
We found that Sea Raft~\citep{wang2025sea}
outperforms Raft~\citep{teed2020raft} on FDS, 
indicating that a better optical flow model 
can lead to more significant improvements.
Additionally, the render quality of RGB 
images shows a slight improvement, 
by 0.58 in 3DGS and 0.50 in 2DGS, 
benefiting from the incorporation of cross-view consistency in FDS. 
On the Mushroom
dataset, adding the FDS loss increases 
the training time by half an hour, which maintains the same
level as baseline.
Similarly, our method shows a notable improvement on the ScanNet dataset as well using Sea Raft~\citep{wang2025sea} Model. The "abs rel" metric in 2DGS is improved nearly 50\%. This demonstrates the robustness and effectiveness of the FDS method across different datasets.
%


The qualitative comparisons on the Mushroom and ScanNet dataset 
are illustrated in \figref{fig:compare}. 
As seen in the first row of \figref{fig:compare}, 
both the original 3DGS and 2DGS suffer from overfitting, 
leading to corrupted geometry generation. 
Our FDS effectively mitigates this issue by 
supervising the matching relationship between 
the input and sampled views, 
helping to recover the geometry.
FDS also improves the refinement of geometric details, 
as shown in other rows. 
By incorporating the matching prior through FDS, 
the quality of the rendered depth is significantly improved.

\begin{table}[t] \centering
\begin{minipage}[t]{0.96\linewidth}
        \captionof{table}{\textbf{3D Reconstruction 
        and novel view synthesis results on Mushroom dataset. * 
        Represents that FDS uses the Raft model.
        }}
        \label{tab:mushroom}
        \resizebox{\textwidth}{!}{
\begin{tabular}{c| c c c c c | c c c c c}
    \hline
     Method &  Acc$\downarrow$ & Comp $\downarrow$ & C-L1 $\downarrow$ & NC $\uparrow$ & F-Score $\uparrow$ &  Abs Rel $\downarrow$ &  PSNR $\uparrow$  & SSIM  $\uparrow$ & LPIPS $\downarrow$ & Time  $\downarrow$ \\ \hline

    GOF &  0.1812 & 0.1093 & 0.1453 & 0.6292 & 0.3665 & 0.2380  & 21.37  &  0.7762  & 0.3132  & $\approx$1.4h\\ 
    PGSR &  0.0971 & 0.1420 & 0.1196 & 0.7193 & 0.5105 & 0.1723  & 22.13  & 0.7773  & 0.2918  & $\approx$1.2h \\ \hline
    3DGS &   0.1167 &  0.1033&  0.1100&  0.7954&  0.3739&  0.1214&  24.18&  0.8392& 0.2511 &$\approx$0.8h \\
    3DGS + FDS$^*$ & 0.0569  & 0.0676 & 0.0623 & 0.8105 & 0.6573 & 0.0603 & 24.72  & 0.8489 & 0.2379 &$\approx$1.3h \\
    3DGS + FDS & \textbf{0.0527}  & \textbf{0.0565} & \textbf{0.0546} & \textbf{0.8178} & \textbf{0.6958} & \textbf{0.0568} & \textbf{24.76}  & \textbf{0.8486} & \textbf{0.2381} &$\approx$1.3h \\ \hline
    2DGS&   0.1078&  0.0850&  0.0964&  0.7835&  0.5170&  0.1002&  23.56&  0.8166& 0.2730 &$\approx$0.8h\\
    2DGS + FDS$^*$ &  0.0689 &  0.0646& 0.0667& 0.8042& 0.6582& 0.0589& 23.98&  0.8255&0.2621 &$\approx$1.3h\\
    2DGS + FDS &  \textbf{0.0615} & \textbf{ 0.0534}& \textbf{0.0574}& \textbf{0.8151}& \textbf{0.6974}&  \textbf{0.0561}&  \textbf{24.06}&  \textbf{0.8271}&\textbf{0.2610} &$\approx$1.3h \\ \hline
\end{tabular}
}
\end{minipage}\hfill
\end{table}

\begin{table}[t] \centering
\begin{minipage}[t]{0.96\linewidth}
        \captionof{table}{\textbf{3D Reconstruction 
        and novel view synthesis results on ScanNet dataset.}}
        \label{tab:scannet}
        \resizebox{\textwidth}{!}{
\begin{tabular}{c| c c c c c | c c c c }
    \hline
     Method &  Acc $\downarrow$ & Comp $\downarrow$ & C-L1 $\downarrow$ & NC $\uparrow$ & F-Score $\uparrow$ &  Abs Rel $\downarrow$ &  PSNR $\uparrow$  & SSIM  $\uparrow$ & LPIPS $\downarrow$ \\ \hline
    GOF & 1.8671  & 0.0805 & 0.9738 & 0.5622 & 0.2526 & 0.1597  & 21.55  & 0.7575  & 0.3881 \\
    PGSR &  0.2928 & 0.5103 & 0.4015 & 0.5567 & 0.1926 & 0.1661  & 21.71 & 0.7699  & 0.3899 \\ \hline

    3DGS &  0.4867 & 0.1211 & 0.3039 & 0.7342& 0.3059 & 0.1227 & 22.19& 0.7837 & 0.3907\\
    3DGS + FDS &  \textbf{0.2458} & \textbf{0.0787} & \textbf{0.1622} & \textbf{0.7831} & 
    \textbf{0.4482} & \textbf{0.0573} & \textbf{22.83} & \textbf{0.7911} & \textbf{0.3826} \\ \hline
    2DGS &  0.2658 & 0.0845 & 0.1752 & 0.7504& 0.4464 & 0.0831 & 22.59& 0.7881 & 0.3854\\
    2DGS + FDS &  \textbf{0.1457} & \textbf{0.0679} & \textbf{0.1068} & \textbf{0.7883} & 
    \textbf{0.5459} & \textbf{0.0432} & \textbf{22.91} & \textbf{0.7928} & \textbf{0.3800} \\ \hline
\end{tabular}
}
\end{minipage}\hfill
\end{table}

\begin{table}[t] \centering
\begin{minipage}[t]{0.96\linewidth}
        \captionof{table}{\textbf{Ablation study on geometry priors.}}
        \label{tab:analysis_prior}
        \resizebox{\textwidth}{!}{
\begin{tabular}{c| c c c c c | c c c c}

    \hline
     Method &  Acc$\downarrow$ & Comp $\downarrow$ & C-L1 $\downarrow$ & NC $\uparrow$ & F-Score $\uparrow$ &  Abs Rel $\downarrow$ &  PSNR $\uparrow$  & SSIM  $\uparrow$ & LPIPS $\downarrow$ \\ \hline
    2DGS&   0.1078&  0.0850&  0.0964&  0.7835&  0.5170&  0.1002&  23.56&  0.8166& 0.2730\\
    2DGS+Depth&   0.0862&  0.0702&  0.0782&  0.8153&  0.5965&  0.0672&  23.92&  0.8227& 0.2619 \\
    2DGS+MVDepth&   0.2065&  0.0917&  0.1491&  0.7832&  0.3178&  0.0792&  23.74&  0.8193& 0.2692 \\
    2DGS+Normal&   0.0939&  0.0637&  0.0788&  \textbf{0.8359}&  0.5782&  0.0768&  23.78&  0.8197& 0.2676 \\
    2DGS+FDS &  \textbf{0.0615} & \textbf{ 0.0534}& \textbf{0.0574}& 0.8151& \textbf{0.6974}&  \textbf{0.0561}&  \textbf{24.06}&  \textbf{0.8271}&\textbf{0.2610} \\ \hline
    2DGS+Depth+FDS &  0.0561 &  0.0519& 0.0540& 0.8295& 0.7282&  0.0454&  \textbf{24.22}& \textbf{0.8291}&\textbf{0.2570} \\
    2DGS+Normal+FDS &  \textbf{0.0529} & \textbf{ 0.0450}& \textbf{0.0490}& \textbf{0.8477}& \textbf{0.7430}&  \textbf{0.0443}&  24.10&  0.8283& 0.2590 \\
    2DGS+Depth+Normal &  0.0695 & 0.0513& 0.0604& 0.8540&0.6723&  0.0523&  24.09&  0.8264&0.2575\\ \hline
    2DGS+Depth+Normal+FDS &  \textbf{0.0506} & \textbf{0.0423}& \textbf{0.0464}& \textbf{0.8598}&\textbf{0.7613}&  \textbf{0.0403}&  \textbf{24.22}& 
    \textbf{0.8300}&\textbf{0.0403}\\
    
\bottomrule
\end{tabular}
}
\end{minipage}\hfill
\end{table}

\subsubsection{Mesh extraction}
To further demonstrate the improvement in geometry quality, 
we applied methods used in ~\citep{turkulainen2024dnsplatter} 
to extract meshes from the input views of optimized 3DGS. 
The comparison results are presented  
in \tabref{tab:mushroom}. 
With the integration of FDS, the mesh quality is significantly enhanced compared to the baseline, featuring fewer floaters and more well-defined shapes.

\subsection{Ablation study}

\textbf{Ablation study on geometry priors:} 
To highlight the advantage of incorporating matching priors, 
we incorporated various types of priors generated by different 
models into 2DGS. These include a monocular depth estimation
model (Depth Anything v2)~\citep{yang2024depth}, a two-view depth estimation 
model (Unimatch)~\citep{xu2023unifying}, 
and a monocular normal estimation model (DSINE)~\citep{bae2024rethinking}.
We adapt the scale and shift-invariant loss in Midas~\citep{birkl2023midas} for
monocular depth supervision and L1 loss for two-view depth supervison.
We use Sea Raft~\citep{wang2025sea} as our default optical flow model.
The comparison results on Mushroom dataset 
are shown in ~\tabref{tab:analysis_prior}.
We observe that the normal prior provides accurate shape information, 
enhancing the geometric quality of the radiance field. 
%
%
The multi-view depth prior, hindered by the limited feature overlap 
between input views, fails to offer reliable geometric 
information. We test average "Abs Rel" of multi-view depth prior
, and the result is 0.19, which performs worse than the "Abs Rel" results 
rendered by original 2DGS.
From the results, it can be seen that depth order information provided by monocular depth improves
reconstruction accuracy. Meanwhile, our FDS achieves the best performance among all the priors, 
and by integrating all
three components, we obtained the optimal results.
\begin{figure}[t] \centering
    \makebox[0.16\textwidth]{\scriptsize RF (16000 iters)}
    \makebox[0.16\textwidth]{\scriptsize RF* (20000 iters)}
    \makebox[0.16\textwidth]{\scriptsize RF (20000 iters)  }
    \makebox[0.16\textwidth]{\scriptsize PF (16000 iters)}
    \makebox[0.16\textwidth]{\scriptsize PF (20000 iters)}

    \includegraphics[width=0.16\textwidth]{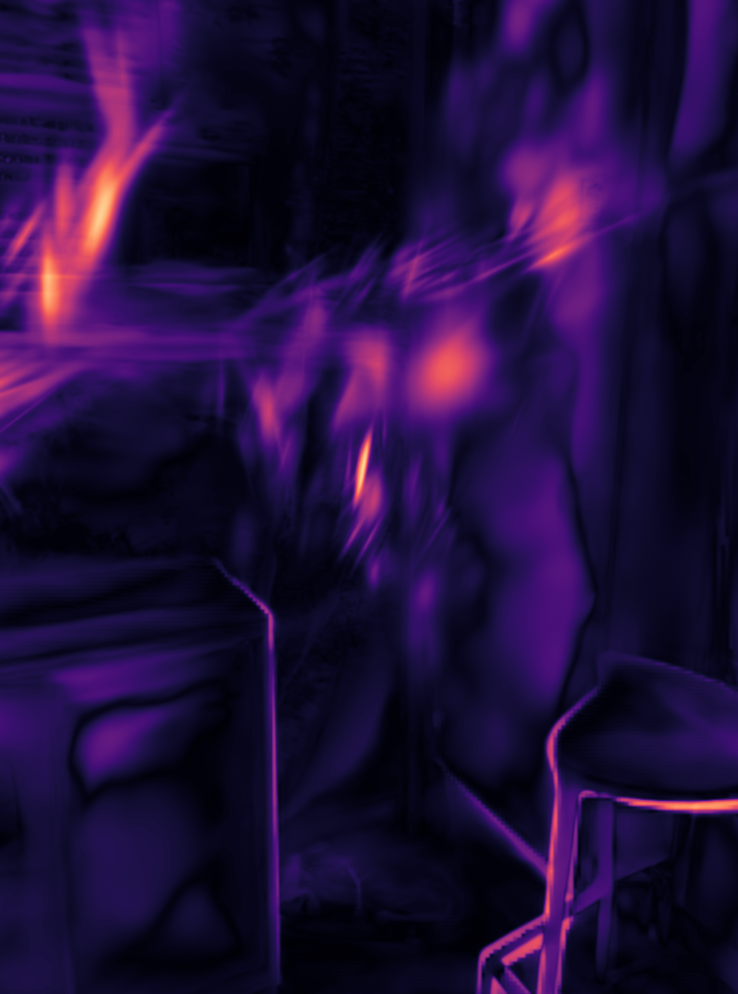}
    \includegraphics[width=0.16\textwidth]{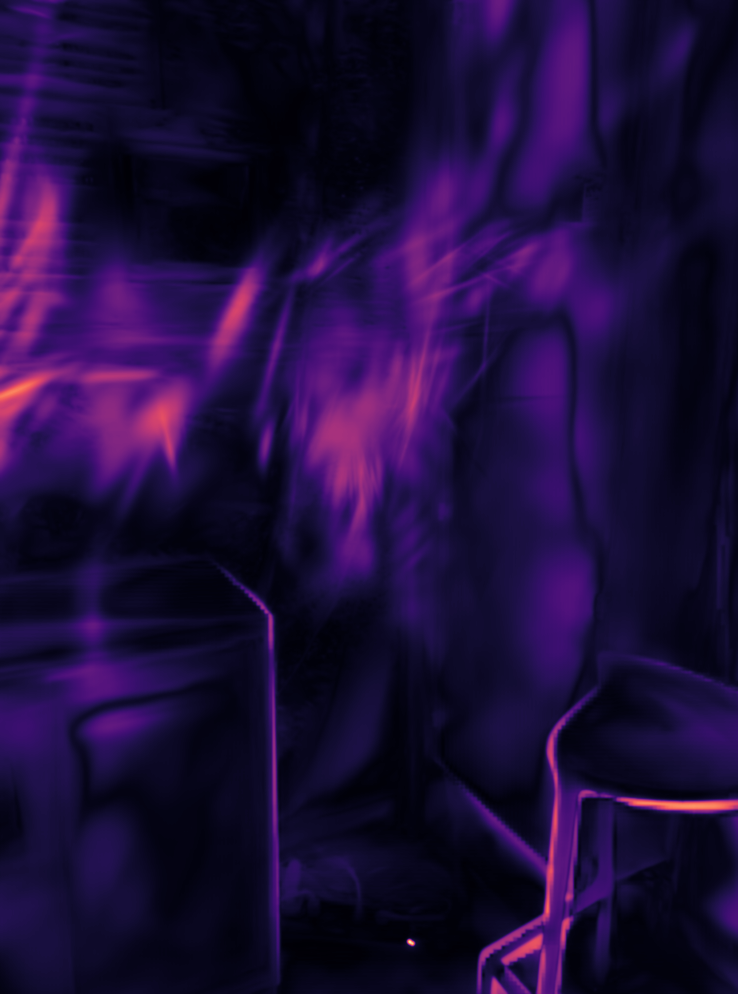}
    \includegraphics[width=0.16\textwidth]{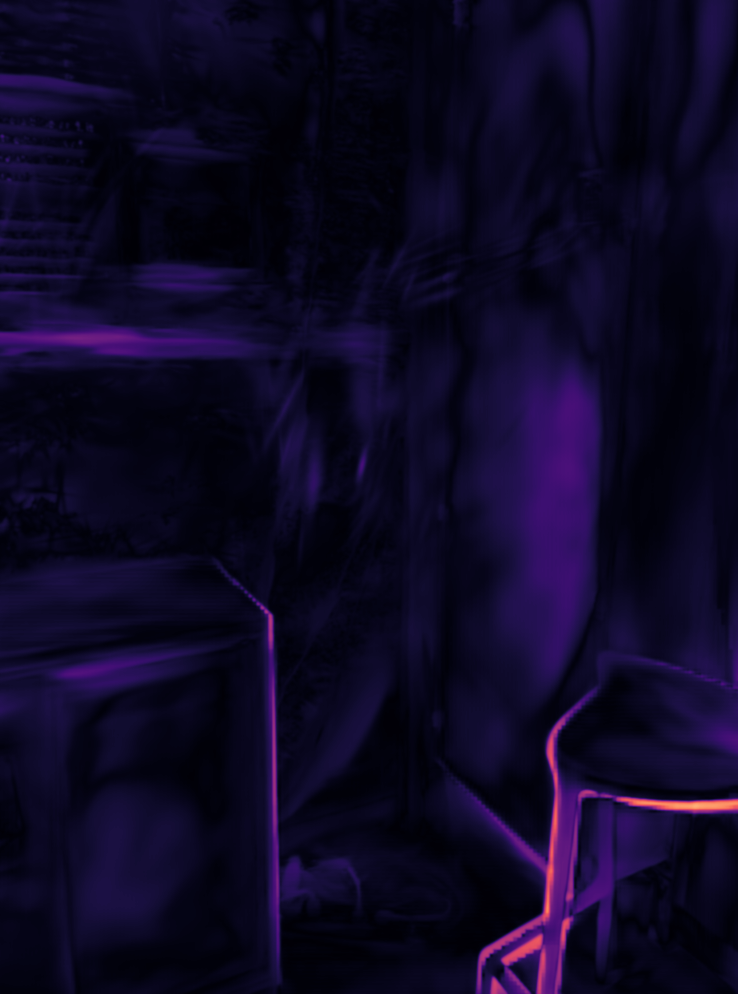}
    \includegraphics[width=0.16\textwidth]{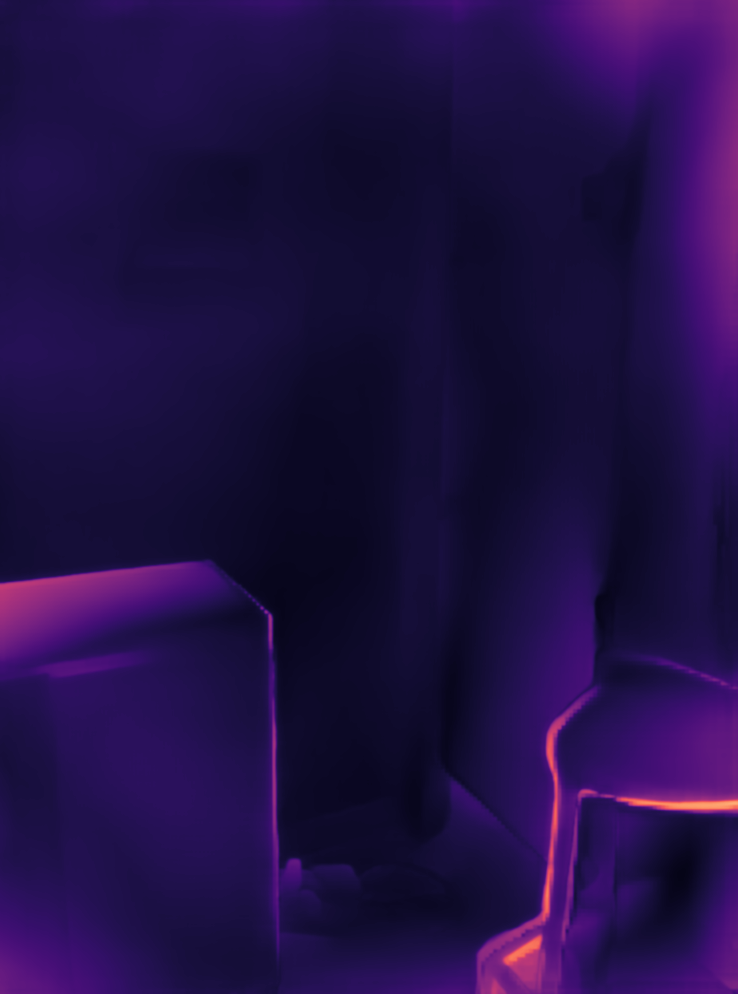}
    \includegraphics[width=0.16\textwidth]{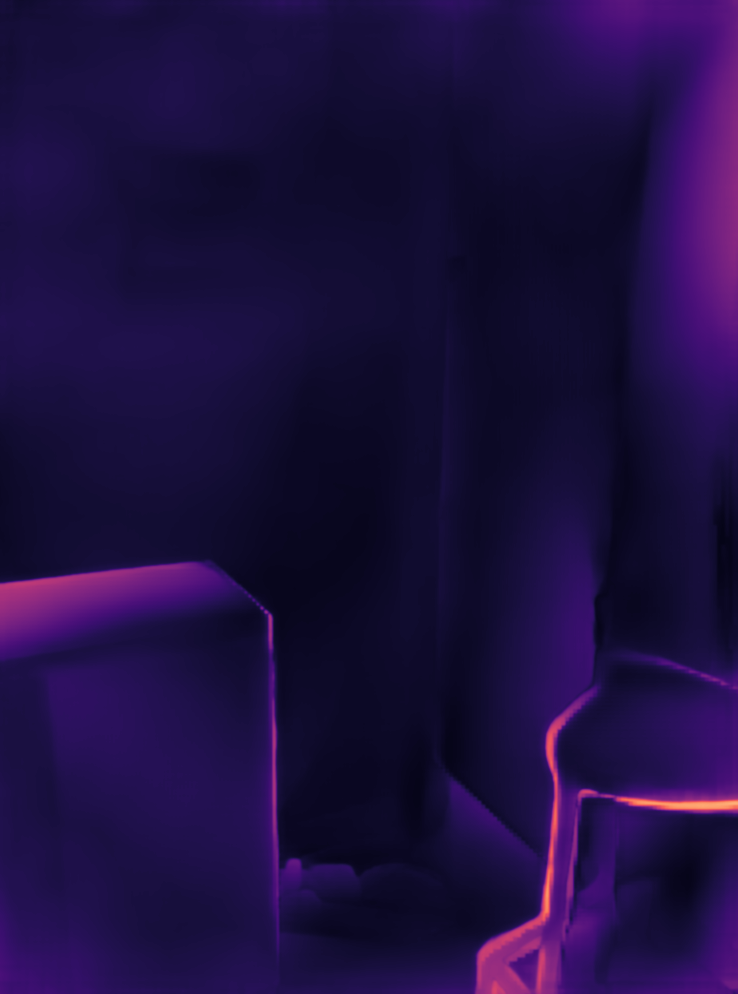}\\
    
    \includegraphics[width=0.16\textwidth]{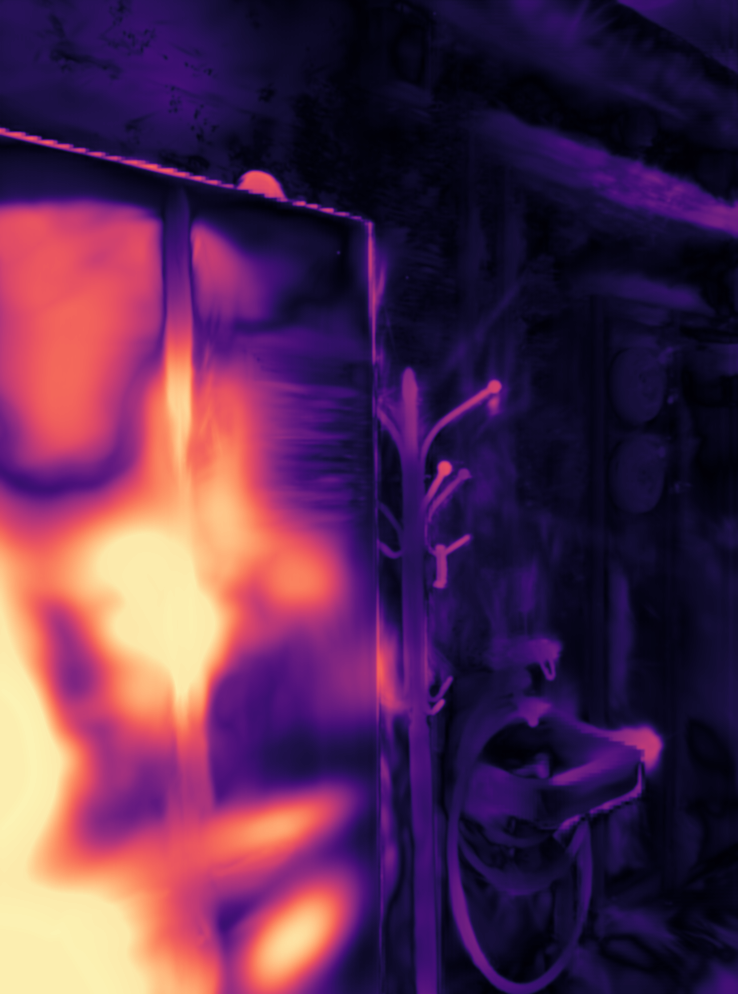}
    \includegraphics[width=0.16\textwidth]{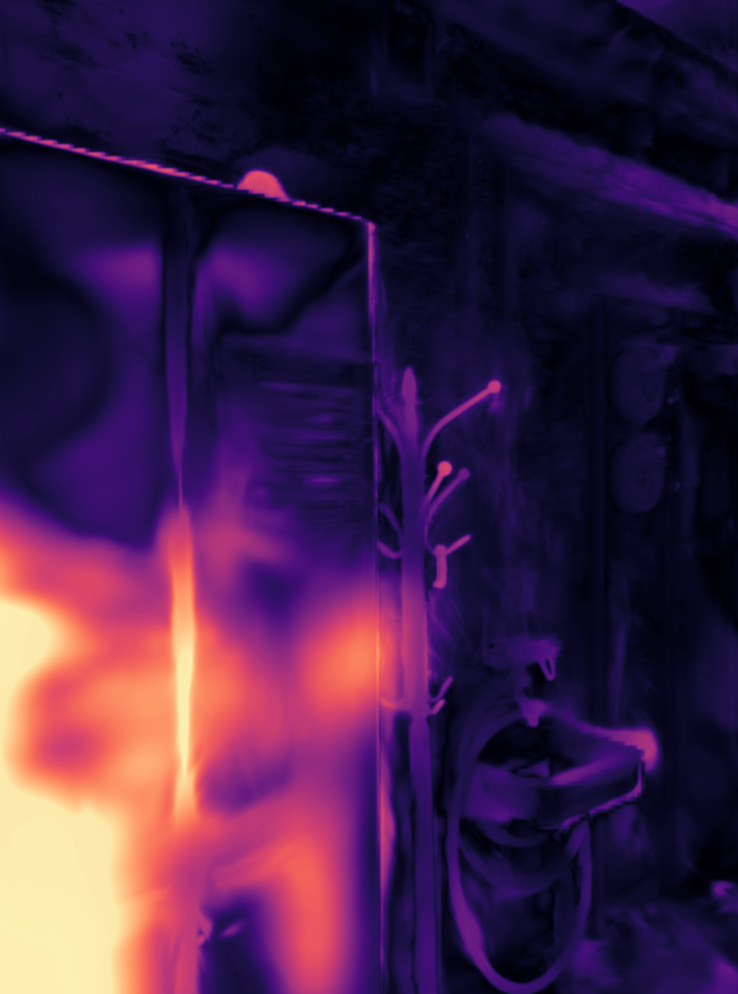}
    \includegraphics[width=0.16\textwidth]{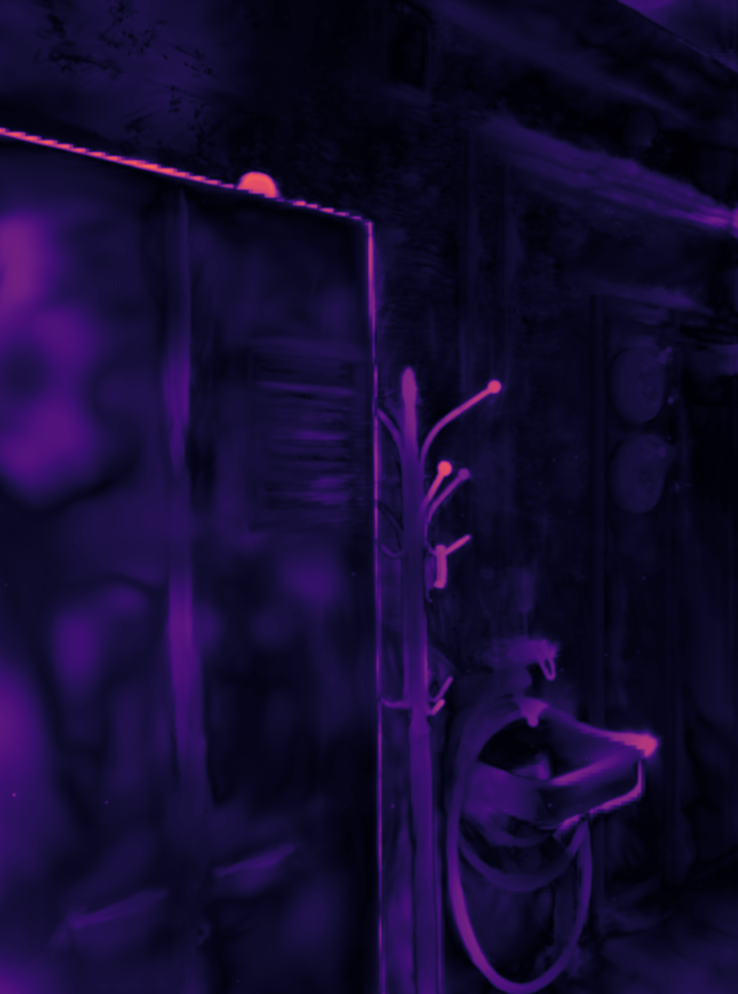}
    \includegraphics[width=0.16\textwidth]{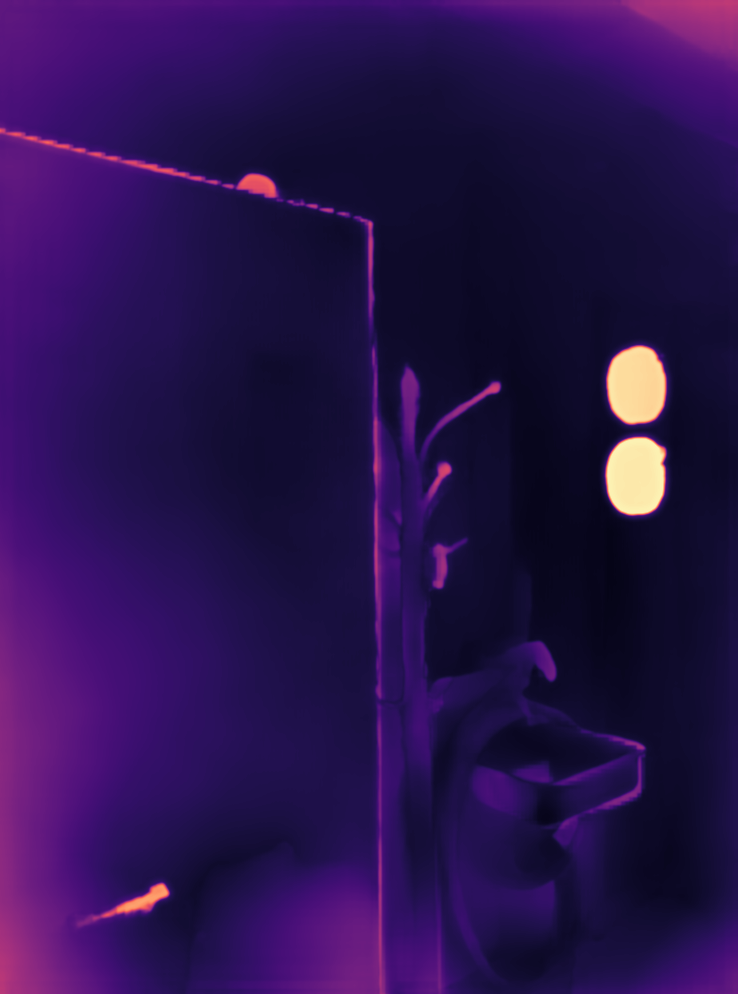}
    \includegraphics[width=0.16\textwidth]{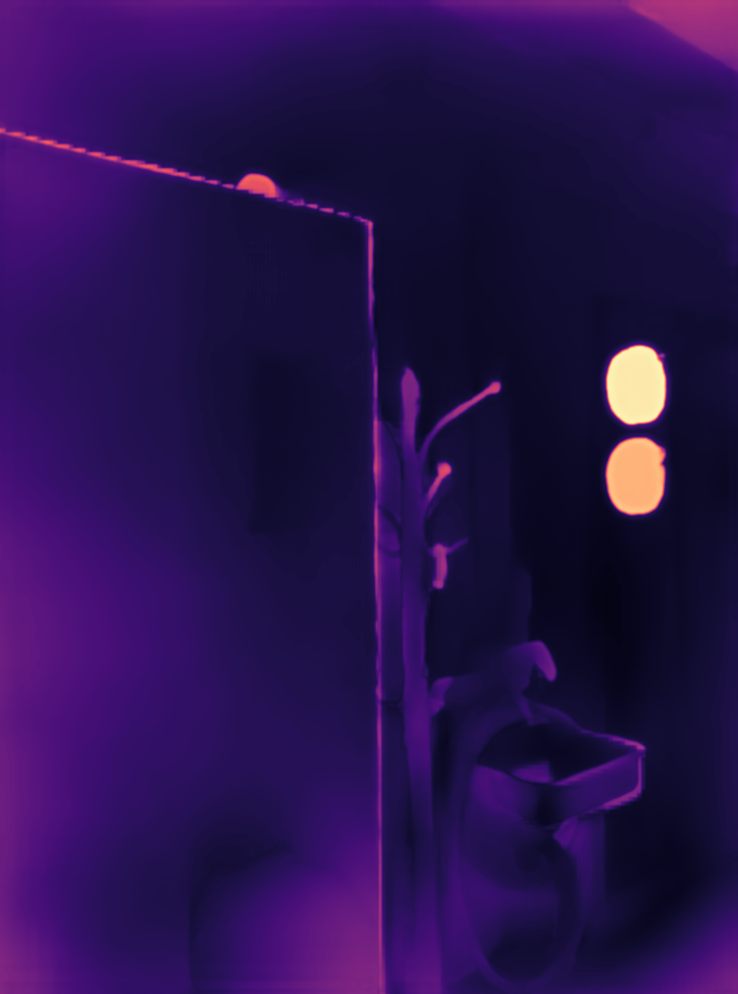}\\

    \includegraphics[width=0.30\textwidth]{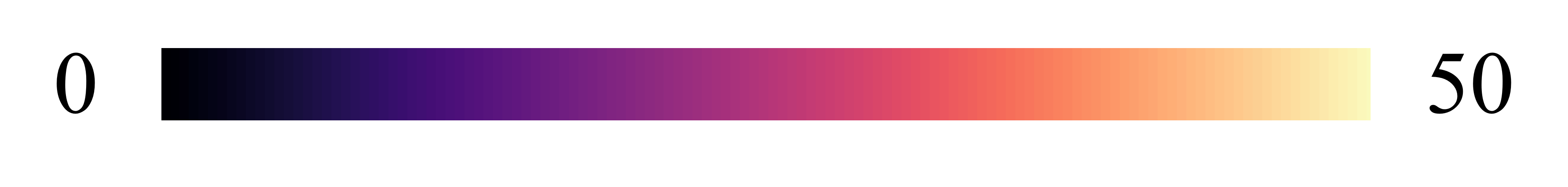}

    \caption{\textbf{The error map of Radiance Flow and Prior Flow.} RF: Radiance Flow, PF: Prior Flow, * means that there is no FDS loss supervision during optimization.}
    \label{fig:error_map}
\end{figure}

\textbf{Ablation study on FDS: }
In this section, we present the design of our FDS 
method through an ablation study on the 
Mushroom dataset to validate its effectiveness.
The optional configurations of FDS are outlined in ~\tabref{tab:ablation_fds}.
Our base model is the 2DGS equipped with FDS,
and its results are shown 
in the first row. The goal of this analysis 
is to evaluate the impact 
of various strategies on FDS sampling and loss design.
We observe that when we 
replace $I_i$ in \eqref{equ:mflow} with $C_i$, 
as shown in the second row, the geometric quality 
of 2DGS deteriorates. Using $I_i$ instead of $C_i$ 
help us to remove the floaters in $\bm{C^s}$, which are also 
remained in $\bm{C^i}$.
We also experiment with modifying the FDS loss. For example, 
in the third row, we use the neighbor 
input view as the sampling view, and replace the 
render result of neighbor view with ground truth image of its input view.
Due to the significant movement between images, the Prior Flow fails to accurately 
match the pixel between them, leading to a further degradation in geometric quality.
Finally, we attempt to fix the sampling view 
and found that this severely damaged the geometric quality, 
indicating that random sampling is essential for the stability 
of the mean error in the Prior flow.

\begin{table}[t] \centering

\begin{minipage}[t]{1.0\linewidth}
        \captionof{table}{\textbf{Ablation study on FDS strategies.}}
        \label{tab:ablation_fds}
        \resizebox{\textwidth}{!}{
\begin{tabular}{c|c|c|c|c|c|c|c}
    \hline
    \multicolumn{2}{c|}{$\mathcal{M}_{\theta}(X, \bm{C^s})$} & \multicolumn{3}{c|}{Loss} & \multicolumn{3}{c}{Metric}  \\
    \hline
    $X=C^i$ & $X=I^i$  & Input view & Sampled view     & Fixed Sampled view        & Abs Rel $\downarrow$ & F-score $\uparrow$ & NC $\uparrow$ \\
    \hline
    & \ding{51} &     &\ding{51}    &    &    \textbf{0.0561}        &  \textbf{0.6974}         & \textbf{0.8151}\\
    \hline
     \ding{51} &           &     &\ding{51}    &    &    0.0839        &  0.6242         &0.8030\\
     &  \ding{51} &   \ding{51}  &    &    &    0.0877       & 0.6091        & 0.7614 \\
      &  \ding{51} &    &    & \ding{51}    &    0.0724           & 0.6312          & 0.8015 \\
\bottomrule
\end{tabular}
}
\end{minipage}
\end{table}

\begin{figure}[htbp] \centering
    \makebox[0.22\textwidth]{}
    \makebox[0.22\textwidth]{}
    \makebox[0.22\textwidth]{}
    \makebox[0.22\textwidth]{}
    \\

    \includegraphics[width=0.22\textwidth]{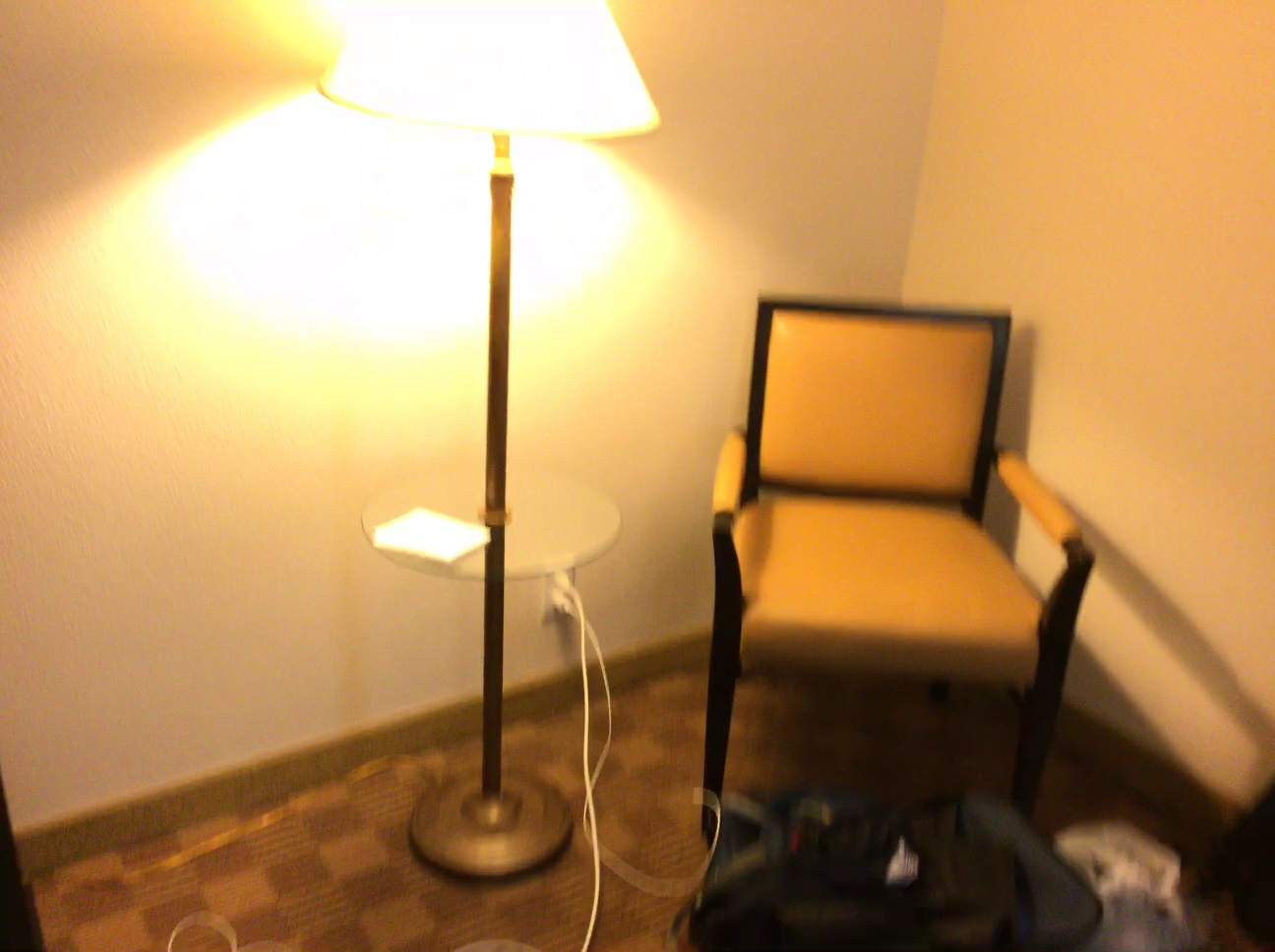}
    \includegraphics[width=0.22\textwidth]{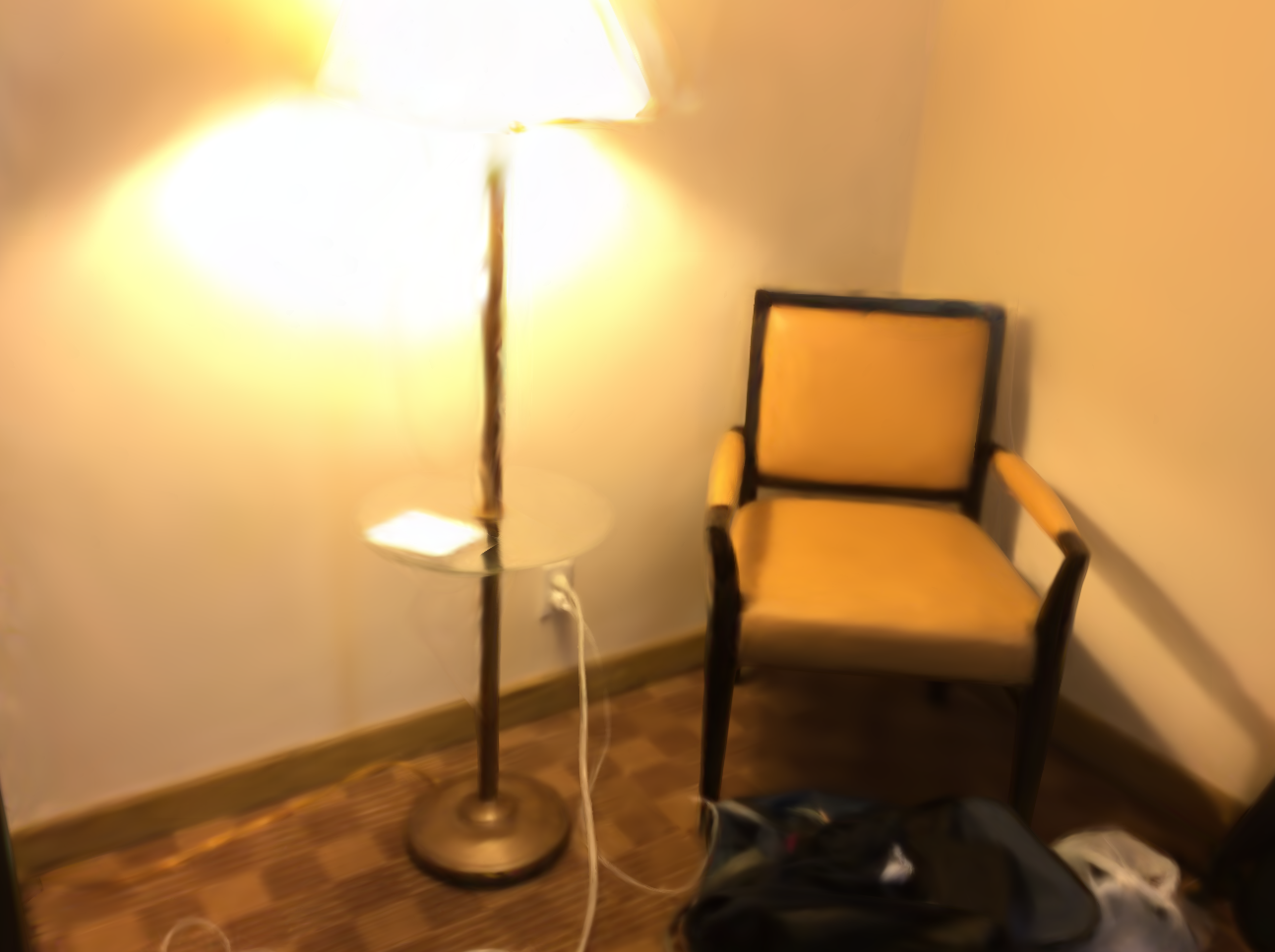}
    \includegraphics[width=0.22\textwidth]{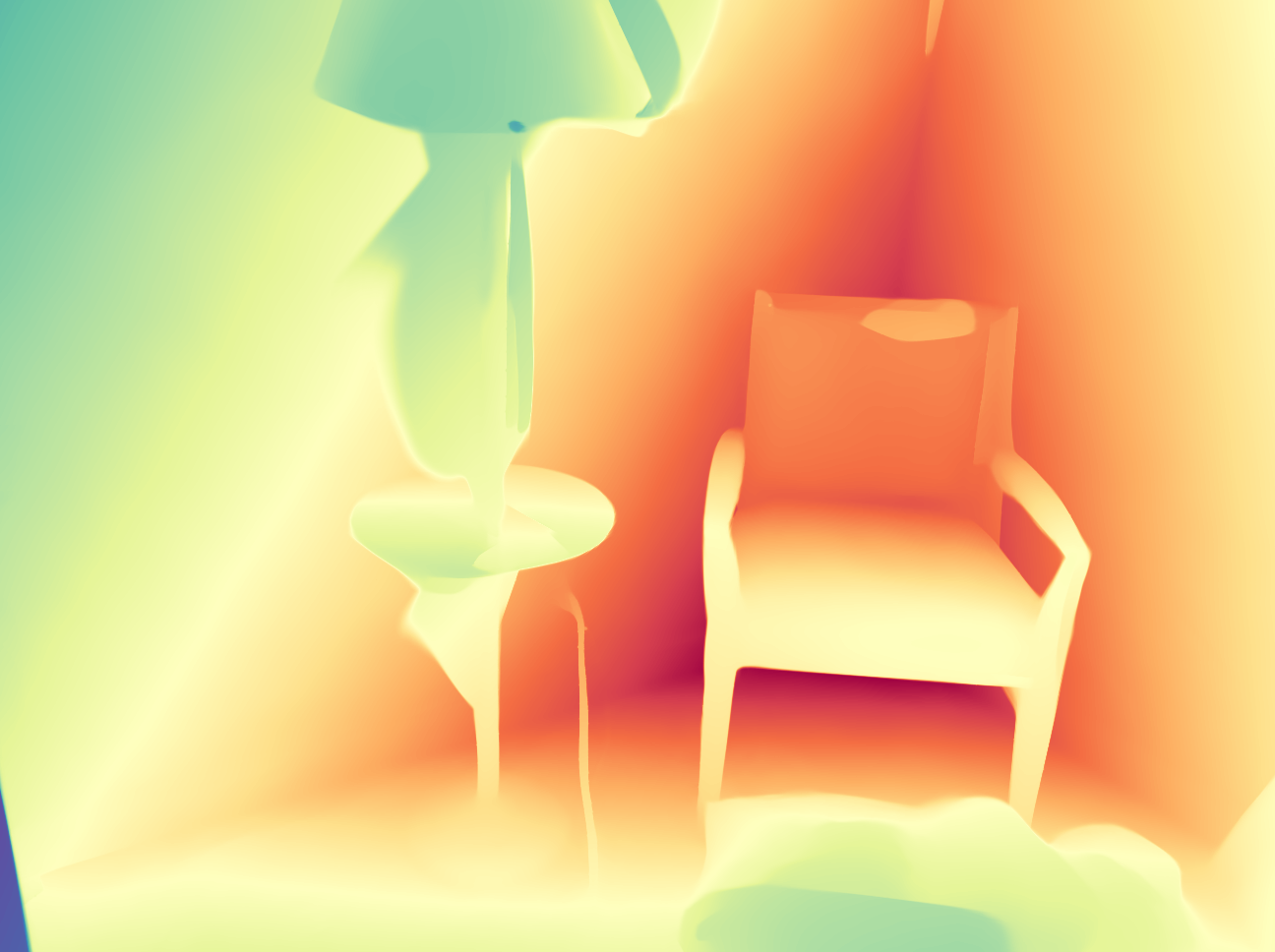}
    \includegraphics[width=0.22\textwidth]{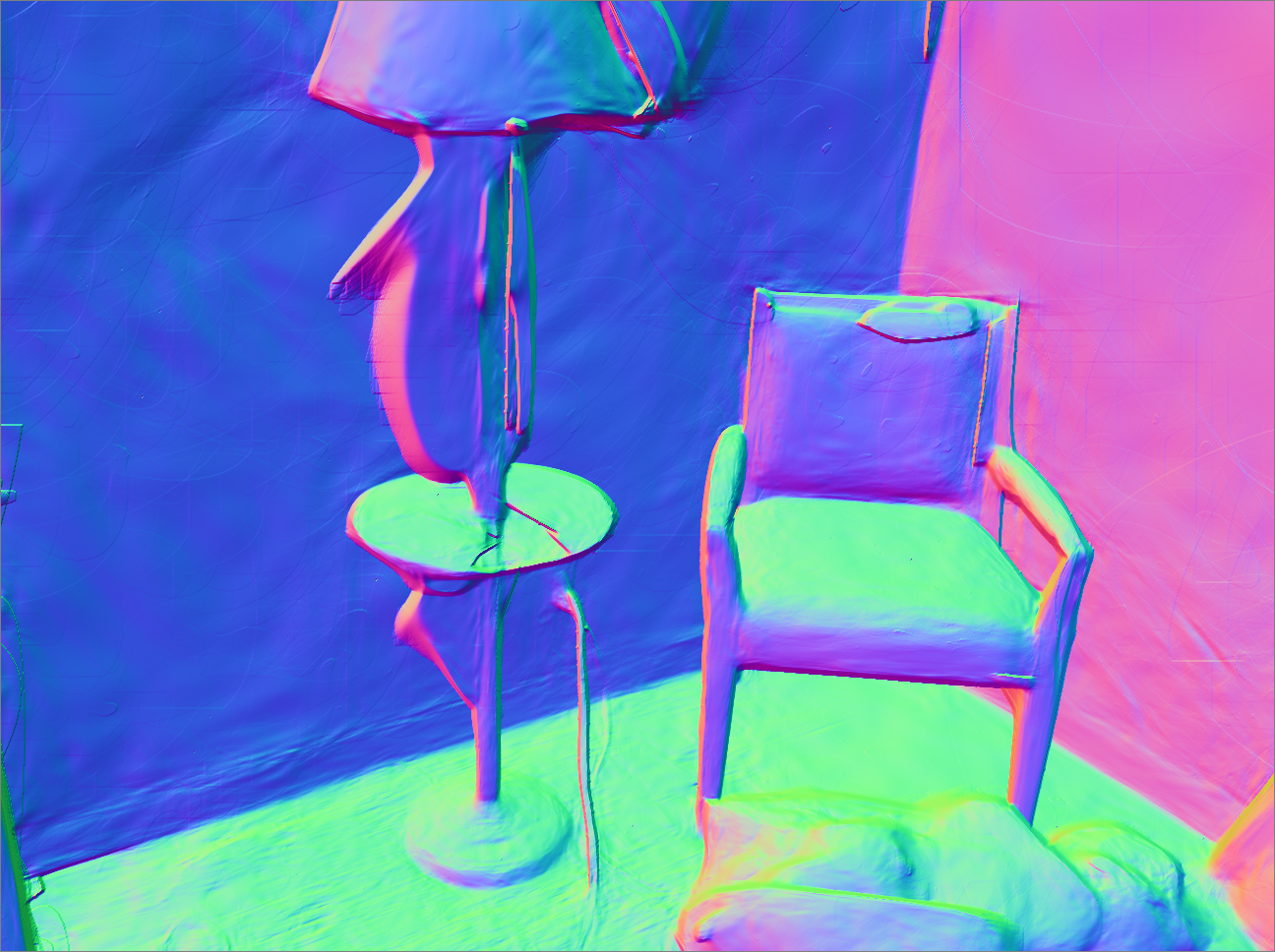}

    \caption{\textbf{Limitation of FDS.} }
    \label{fig:limitation}
\end{figure}


\textbf{Interpretive Experiments: }
To demonstrate the mutual refinement of two flows in our FDS, 
For each view, we sample the unobserved 
views multiple times to compute the mean error 
of both Radiance Flow and Prior Flow. 
We use Raft~\citep{teed2020raft} as our default optical flow model
for visualization.
The ground truth flow is calculated based on 
~\eref{equ:flow_pose} and ~\eref{equ:flow} 
utilizing ground truth depth in dataset.
We introduce our FDS loss after 16000 iterations during 
optimization of 2DGS.
The error maps are shown in ~\figref{fig:error_map}.
Our analysis reveals that Radiance Flow tends to 
exhibit significant geometric errors, 
whereas Prior Flow can more accurately estimate the geometry,
effectively disregarding errors introduced by floating Gaussian points.

\subsection{Limitation and further work}

Firstly, our FDS faces challenges in scenes with 
significant lighting variations between different 
views, as shown in the lamp of first row in ~\figref{fig:limitation}. 
Incorporating exposure compensation into FDS could help address this issue. 
 Additionally, our method struggles with 
 reflective surfaces and motion blur,
 leading to incorrect matching. 
 In the future, we plan to explore the potential 
 of FDS in monocular video reconstruction tasks, 
 using only a single input image at each time step.

\section{Conclusions}
In this paper, we propose Flow Distillation Sampling (FDS), which
leverages the matching prior between input views and 
sampled unobserved views from the pretrained optical flow model, to improve the geometry quality
of Gaussian radiance field. 
Our method can be applied to different approaches (3DGS and 2DGS) to enhance the geometric rendering quality of the corresponding neural radiance fields.
We apply our method to the 3DGS-based framework, 
and the geometry is enhanced on the Mushroom, ScanNet, and Replica datasets.

\section*{Acknowledgements} This work was supported by 
National Key R\&D Program of China (2023YFB3209702), 
the National Natural Science Foundation of 
China (62441204, 62472213), and Gusu 
Innovation \& Entrepreneurship Leading Talents Program (ZXL2024361)
\bibliographystyle{plainnat}
\bibliography{egbib}
\end{document}